\documentclass{article}

\usepackage{microtype}
\usepackage{graphicx}
\usepackage{subfigure}
\usepackage{booktabs}

\usepackage{hyperref}

\usepackage[accepted]{icml2024}

\usepackage{amsmath}
\usepackage{amssymb}
\usepackage{mathtools}
\usepackage{amsthm}

\usepackage[capitalize,noabbrev]{cleveref}

\theoremstyle{plain}
\newtheorem{theorem}{Theorem}[section]

\theoremstyle{definition}
\newtheorem{definition}[theorem]{Definition}
\newtheorem{assumption}[theorem]{Assumption}
\theoremstyle{remark}

\usepackage[textsize=tiny]{todonotes}

\usepackage{amsthm}

\usepackage{centernot}
\usepackage{nicefrac}       
\usepackage{mathtools}
\usepackage{amsbsy}
\usepackage{amstext}
\usepackage{thmtools}
\usepackage{thm-restate}

\begingroup
    \makeatletter
    \@for\theoremstyle:=definition,remark,plain\do{%
        \expandafter\g@addto@macro\csname th@\theoremstyle\endcsname{%
            \addtolength\thm@preskip\parskip
            }%
        }
\endgroup

\renewcommand{\mid}{~\vert~}

\usepackage{booktabs,arydshln}
\makeatletter
\def\adl@drawiv#1#2#3{%
        \hskip.5\tabcolsep
        \xleaders#3{#2.5\@tempdimb #1{1}#2.5\@tempdimb}%
                #2\z@ plus1fil minus1fil\relax
        \hskip.5\tabcolsep}
\newcommand{\cdashlinelr}[1]{%
  \noalign{\vskip\aboverulesep
           \global\let\@dashdrawstore\adl@draw
           \global\let\adl@draw\adl@drawiv}
  \cdashline{#1}
  \noalign{\global\let\adl@draw\@dashdrawstore
           \vskip\belowrulesep}}
\makeatother

\renewcommand{\epsilon}{\varepsilon}

\newenvironment{example*}
 {\pushQED{\qed}\example}
 {\popQED\endexample}
\numberwithin{equation}{section}

\newcommand{\defnphrase}[1]{\emph{#1}}

\newcommand{\Reals}{\mathbb{R}}

\DeclareMathOperator*{\logit}{logit}

\renewcommand{\Pr}{\mathbb{P}}

\newcommand{\given}{\mid}

\providecommand\given{} 
\newcommand\SetSymbol[1][]{
  \nonscript\,#1:\nonscript\,\mathopen{}\allowbreak}
\DeclarePairedDelimiterX\Set[1]{\lbrace}{\rbrace}%
{ \renewcommand\given{\SetSymbol[]} #1 }

\usepackage{enumitem}

\usepackage{thm-restate}

\crefformat{equation}{(#2#1#3)}
\crefformat{figure}{Figure~#2#1#3}
\crefname{definition}{Definition}{Definitions}
\crefname{example}{Example}{Examples}
\crefname{lemma}{Lemma}{Lemmas}
\crefname{cor}{Corollary}{Corollaries}
\crefname{theorem}{Theorem}{Theorems}
\crefname{assumption}{Assumption}{Assumptions}

\DeclareMathOperator{\repOp}{Rep}
\newcommand{\repText}[1]{\repOp(\text{``#1''})}

\newcommand{\ConceptName}[1]{$\mathtt{#1}$}

\newcommand{\ConceptDirName}[2]{\texttt{#1$\Rightarrow$#2}}

\DeclareMathOperator{\ConeOp}{Cone}

\newcommand{\Cone}[1]{\ConeOp(#1)}
\newcommand{\cov}{\mathrm{Cov}}

\newcommand{\ip}[2]{\langle #1,#2\rangle}

\usepackage{tikz}
\usetikzlibrary{positioning, shadows, arrows.meta, bending}

\icmltitlerunning{The Linear Representation Hypothesis and the Geometry of Large Language Models}

\begin{document}

\twocolumn[
\icmltitle{The Linear Representation Hypothesis and\\
            the Geometry of Large Language Models}

\icmlsetsymbol{equal}{*}

\begin{icmlauthorlist}
\icmlauthor{Kiho Park}{uchicago}
\icmlauthor{Yo Joong Choe}{uchicago}
\icmlauthor{Victor Veitch}{uchicago}
\end{icmlauthorlist}

\icmlaffiliation{uchicago}{University of Chicago, Illinois, USA}

\icmlcorrespondingauthor{Kiho Park}{parkkiho@uchicago.edu}
\icmlkeywords{Linear Representation Hypothesis, Causal Inner Product, Interpretability, Machine Learning, ICML}

\vskip 0.3in
]

\printAffiliationsAndNotice{}

\begin{abstract}
Informally, the ``linear representation hypothesis'' is the idea that high-level concepts are represented linearly as directions in some representation space.
In this paper, we address two closely related questions: What does ``linear representation'' actually mean? And, how do we make sense of geometric notions (e.g., cosine similarity and projection) in the representation space?  
To answer these, we use the language of counterfactuals to give two formalizations of linear representation, one in the output (word) representation space, and one in the input (context) space.
We then prove that these connect to linear probing and model steering, respectively.
To make sense of geometric notions, we use the formalization to identify a particular (non-Euclidean) inner product that respects language structure in a sense we make precise.
Using this \emph{causal inner product}, we show how to unify all notions of linear representation.
In particular, this allows the construction of probes and steering vectors using counterfactual pairs.
Experiments with LLaMA-2 demonstrate the existence of linear representations of concepts, the connection to interpretation and control, and the fundamental role of the choice of inner product.
Code is available at \href{https://github.com/KihoPark/linear_rep_geometry}{github.com/KihoPark/linear\_rep\_geometry}.
\end{abstract}

\section{Introduction}
In the context of language models, the ``Linear Representation Hypothesis'' is the idea that high-level concepts are represented linearly in the representation space of a model~\citep[e.g.,][]{mikolov2013linguistic,arora2016latent,elhage2022toy}.
High-level concepts might include: is the text in French or English? Is it in the present or past tense? If the text is about a person, are they male or female?
The appeal of the \emph{linear} representation hypothesis is that---were it true---the tasks of interpreting and controlling model behavior could exploit linear algebraic operations on the representation space.
Our goal is to formalize the linear representation hypothesis,
and clarify how it relates to interpretation and control.

\begin{figure*}[t]
    \centering
    \includegraphics[trim={4.5cm 4.5cm 4.5cm 4.5cm}, clip, width=0.75\linewidth]{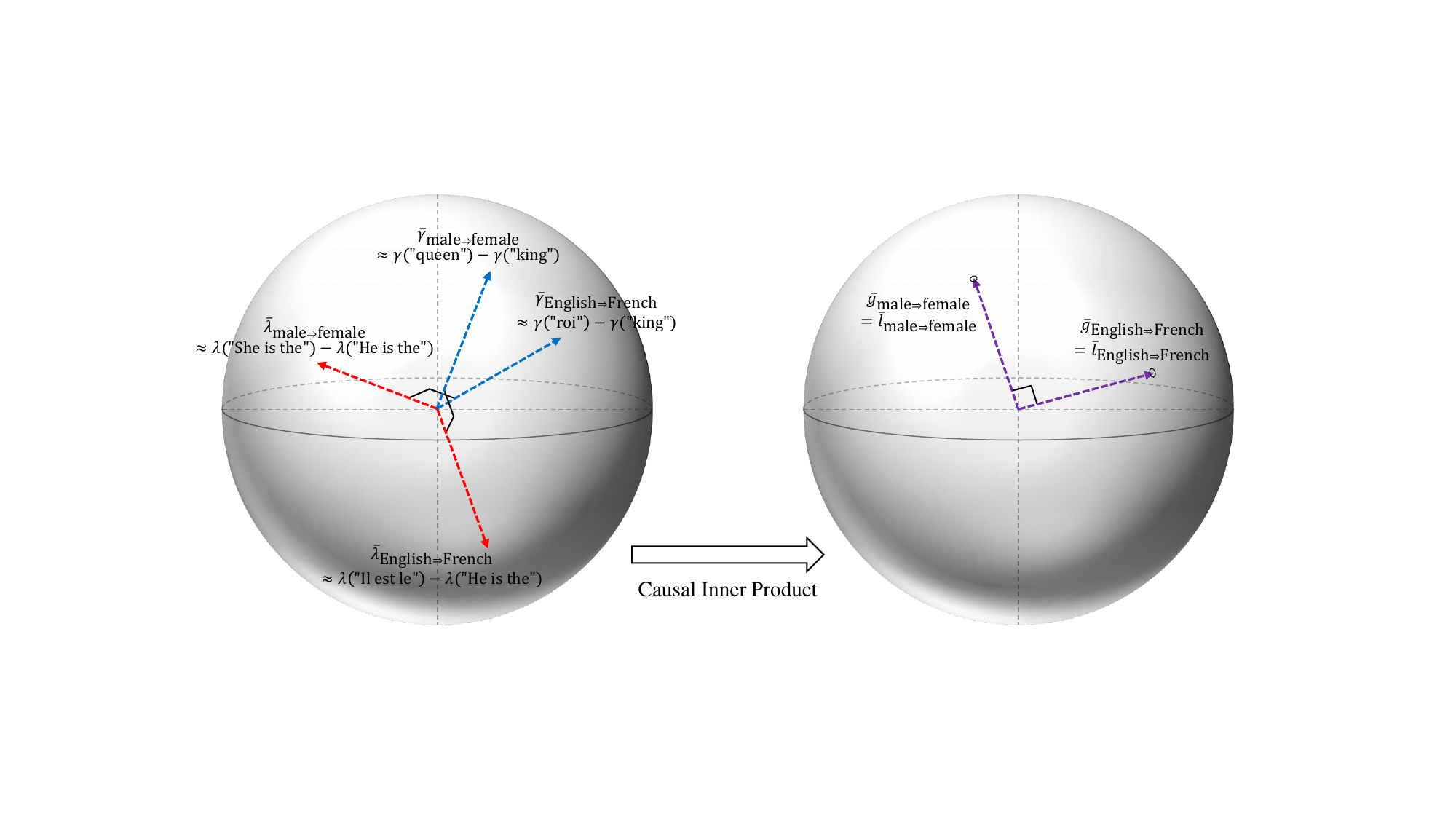}
    \caption{
    The geometry of linear representations can be understood in terms of a \emph{causal inner product} that respects the semantic structure of concepts.
    In a language model, each concept has two separate linear representations, $\bar{\lambda}$ (red) in the embedding (input context) space and $\bar{\gamma}$ (blue) in the unembedding (output word) space, as drawn on the left.
    The causal inner product induces a linear transformation for the representation spaces such that the transformed linear representations coincide (purple), as drawn on the right.
    In this unified representation space, causally separable concepts are represented by orthogonal vectors.
    }
    \label{fig:sphere}
\end{figure*}

The first challenge is that it is not clear what ``linear representation'' means. 
There are (at least) three interpretations:
\begin{enumerate}[topsep=0pt, parsep=0pt, partopsep=0pt]
    \item \textbf{Subspace:} \Citep[e.g.,][]{mikolov2013linguistic,pennington2014glove} The first idea is that each concept is represented as a (1-dimensional) subspace.
    For example, in the context of word embeddings, it has been argued empirically that $\repText{woman}-\repText{man}$, $\repText{queen}-\repText{king}$, and all similar pairs belong to a common subspace \citep{mikolov2013linguistic}. Then, it is natural to take this subspace to be a representation of the concept of \ConceptName{Male/Female}.
    \item \textbf{Measurement:} \Citep[e.g.,][]{nanda2023emergent,LMSpaceTime:2023} Next is the idea that the probability of a concept value can be measured with a linear probe.
    For example, the probability that the output language is French is logit-linear in the representation of the input.
    In this case, we can take the linear map to be a representation of the concept of \ConceptName{English/French}.
    \item \textbf{Intervention:}  \Citep[e.g.,][]{wang2023concept,ActivationAddition:2023} The final idea is that the value a concept takes on can be changed, without changing other concepts, by adding a suitable steering vector---e.g., we change the output from English to French by adding an \ConceptName{English/French} vector. In this case, we take this added vector to be a representation of the concept.
\end{enumerate}
It is not clear a priori how these ideas relate to each other, nor which is the ``right'' notion of linear representation.

Next, suppose we have somehow found the linear representations of various concepts. We can then use linear algebraic operations on the representation space for interpretation and control. For example, we might compute the cosine similarity between a representation and known concept directions, or edit representations projected onto target directions.
However, similarity and projection are geometric notions: they require an inner product on the representation space. The second challenge is that it is not clear which inner product is appropriate for understanding model representations.

To address these, we make the following contributions:
\begin{enumerate}[topsep=0pt, parsep=0pt, partopsep=0pt]
    \item First, we formalize the subspace notion of linear representation in terms of counterfactual pairs, in both ``embedding'' (input context) and ``unembedding'' (output word) spaces. Using this formalization, we prove that the unembedding notion connects to measurement, and the embedding notion to intervention.
    \item Next, we introduce the notion of a \emph{causal inner product}: an inner product with the property that concepts that can vary freely of each other are represented as orthogonal vectors. 
    We show that such an inner product has the special property that it unifies the embedding and unembedding representations, as illustrated in \cref{fig:sphere}.
    Additionally, we show how to estimate the inner product using the LLM unembedding matrix.
    \item Finally, we study the linear representation hypothesis empirically using LLaMA-2 \citep{touvron2023llama2}.
    We find the subspace notion of linear representations for a variety of concepts.
    Using these, we give evidence that the causal inner product respects semantic structure, and that subspace representations can be used to construct measurement and intervention representations.
\end{enumerate}

\paragraph{Background on Language Models}
We will require some minimal background on (large) language models.
Formally, a language model takes in context text $x$ and samples output text.
This sampling is done word by word (or token by token).
Accordingly, we'll view the outputs as single words.
To define a probability distribution over outputs, the language model first maps each context $x$ to a vector $\lambda(x)$ in a representation space $\Lambda \simeq \Reals^d$. We will call these \defnphrase{embedding vectors}.
The model also represents each word $y$ as an \defnphrase{unembedding vector} $\gamma(y)$ in a separate representation space $\Gamma \simeq \Reals^d$.
The probability distribution over the next words is then given by the softmax distribution:
\begin{equation*}
    \Pr(y \given x) \propto \exp(\lambda(x)^\top \gamma(y)).
\end{equation*}

\section{The Linear Representation Hypothesis}\label{sec:linear_rep_hyp}
We begin by formalizing the subspace notion of linear representation, one in each of the unembedding and embedding spaces of language models, and then tie the subspace notions to the measurement and intervention notions.

\subsection{Concepts}
The first step is to formalize the notion of a concept. Intuitively, a concept is any factor of variation that can be changed in isolation. For example, we can change the output from French to English without changing its meaning, or change the output from being about a man to about a woman without changing the language it is written in.

Following \citet{wang2023concept}, we formalize this idea by taking a \emph{concept variable} $W$ to be a latent variable that is caused by the context $X$, and that acts as a cause of the output $Y$.
For simplicity of exposition, we will restrict attention to binary concepts.
Anticipating the representation of concepts by vectors, we introduce an ordering on each binary concept---e.g., \ConceptDirName{male}{female}. This ordering makes the sign of a representation meaningful (e.g., the representation of \ConceptDirName{female}{male} will have the opposite sign).

Each concept variable $W$ defines a set of counterfactual outputs $\{Y(W=w)\}$ that differ only in the value of $W$.
For example, for the concept \ConceptDirName{male}{female}, $(Y(0),Y(1))$ is a random element of the set \{(``man'', ``woman''), (``king'', ``queen''), \dots\}. In this paper, we assume the value of concepts can be read off deterministically from the sampled output (e.g., the output ``king'' implies $W=0$). Then, we can specify concepts by specifying their corresponding counterfactual outputs.

We will eventually need to reason about the relationships between multiple concepts. 
We say that two concepts $W$ and $Z$ are \emph{causally separable} if $Y(W=w,Z=z)$ is well-defined for each $w,z$. That is, causally separable concepts are those that can be varied freely and in isolation.
For example, the concepts \ConceptDirName{English}{French} and \ConceptDirName{male}{female} are causally separable---consider $\{\text{``king''}, \text{``queen''}, \text{``roi''}, \text{``reine''}\}$.
However, the concepts \ConceptDirName{English}{French} and \ConceptDirName{English}{Russian} are not because they cannot vary freely.

We'll write $Y(W=w, Z=z)$ as $Y(w,z)$ when the concepts are clear from context.

\subsection{Unembedding Representations and Measurement}
We now turn to formalizing linear representations of a concept.
The first observation is that there are two distinct representation spaces in play---the embedding space $\Lambda$ and the unembedding space $\Gamma$.
A concept could be linearly represented in either space.
We begin with the unembedding space.
Defining the cone of vector $v$ as $\Cone{v} = \{\alpha v: \alpha > 0\}$,
\begin{definition}[Unembedding Representation]\label{def:unembedding_rep}
    We say that $\bar{\gamma}_W$ is an \defnphrase{unembedding representation} of a concept $W$ if $\gamma(Y(1)) - \gamma(Y(0)) \in \Cone{\bar{\gamma}_W}$ almost surely.
\end{definition}
This definition captures the subspace notion in the unembedding space, e.g., that $\gamma(\text{``queen''}) - \gamma(\text{``king''})$ is parallel to $\gamma(\text{``woman''}) - \gamma(\text{``man''})$.
We use a cone instead of subspace because the sign of the difference is significant---i.e., the difference between ``king'' and ``queen'' is in the opposite direction as the difference between ``woman'' and ``man''.
The unembedding representation (if it exists) is unique up to positive scaling, consistent with the linear subspace hypothesis that concepts are represented as directions.

\paragraph{Connection to Measurement}
The first result is that the unembedding representation is closely tied to the measurement notion of linear representation:
\begin{restatable}[Measurement Representation]{theorem}{measRep}\label{thm:measurement}
    Let $W$ be a concept, and let $\bar\gamma_W$ be the unembedding representation of $W$. Then, given any context embedding $\lambda \in \Lambda$,
    \begin{align*}
        \logit \Pr(Y=Y(1) \given Y\in\{Y(0),Y(1)\}, \lambda)
        = \alpha \lambda^\top \bar{\gamma}_W,
    \end{align*}
    where $\alpha > 0$ (a.s.) is a function of $\{Y(0),Y(1)\}$. 
\end{restatable}
All proofs are given in \Cref{sec:proofs}.

In words: if we know the output token is either ``king'' or ``queen'' (say, the context was about a monarch), then the probability that the output is ``king'' is logit-linear in the language model representation with regression coefficients $\bar{\gamma}_W$. 
The random scalar $\alpha$ is a function of the particular counterfactual pair $\{Y(0),Y(1)\}$---e.g., it may be different for $\{\text{``king''}, \text{``queen''}\}$ and $\{\text{``roi''}, \text{``reine''}\}$. However, the direction used for prediction is the same for all counterfactual pairs demonstrating the concept.

\Cref{thm:measurement} shows a connection between the subspace representation and the linear representation learned by fitting a linear probe to predict the concept.
Namely, in both cases, we get a predictor that is linear on the logit scale. 
However, the unembedding representation differs from a probe-based representation in that it does not incorporate any information about correlated but off-target concepts. For example, if French text were disproportionately about men, a probe could learn this information (and include it in the representation), but the unembedding representation would not.
In this sense, the unembedding representation might be viewed as an ideal probing representation.

\subsection{Embedding Representations and Intervention}
The next step is to define a linear subspace representation in the embedding space $\Lambda$.
We'll again go with a notion anchored in demonstrative pairs.
In the embedding space, each $\lambda(x)$ defines a distribution over concepts.
We consider pairs of sentences such as $\lambda_0 = \lambda(\text{``He is the monarch of England, ''})$ and $\lambda_1 = \lambda(\text{``She is the monarch of England, ''})$ that induce different distributions on the target concept, but the same distribution on all off-target concepts. A concept is embedding-represented if the differences between all such pairs belong to a common subspace. Formally,
\begin{definition}
    [Embedding Representation]\label{def:embedding_rep}
    We say that $\bar{\lambda}_W$ is an \defnphrase{embedding representation} of a concept $W$ if we have
    $\lambda_1 - \lambda_0 \in \Cone{\bar{\lambda}_W}$ for any context embeddings $\lambda_0, \lambda_1 \in \Lambda$ that satisfy
    \begin{equation*}
        \frac{\Pr(W=1 \given \lambda_1)}{\Pr(W=1\given \lambda_0)} > 1 \quad\text{and}\quad \frac{\Pr(W,Z \given \lambda_1)}{\Pr(W, Z \given \lambda_0)} = \frac{\Pr(W \given \lambda_1)}{\Pr(W \given \lambda_0)},
    \end{equation*}
    for each concept $Z$ that is causally separable with $W$.
\end{definition} 
The first condition ensures that the direction is relevant to the target concept, and the second condition ensures that the direction is not relevant to off-target concepts.

\paragraph{Connection to Intervention}
It turns out the embedding representation is closely tied to the intervention notion of linear representation.
For this, we need the following lemma relating embedding and unembedding representations.
\begin{restatable}[Unembedding-Embedding Relationship]{lemma}{UnembeddingEmbedding}\label{lem:unembedding-embedding}
    Let $\bar{\lambda}_W$ be the embedding representation of a concept $W$, and let $\bar\gamma_W$ and $\bar\gamma_Z$ be the unembedding representations for $W$ and any concept $Z$ that is causally separable with $W$. Then,
    \begin{equation}\label{eq:lemma}
        \bar{\lambda}_W^\top \bar{\gamma}_W > 0 \quad \text{and}\quad \bar{\lambda}_W^\top \bar{\gamma}_Z = 0 .
    \end{equation}
    Conversely, if a representation $\bar\lambda_W$ satisfies \eqref{eq:lemma}, and if there exist concepts $\{Z_i\}_{i=1}^{d-1}$, such that each $Z_i$ is causally separable with $W$ and $\{\bar\gamma_W\}\cup\{\bar\gamma_{Z_i}\}_{i=1}^{d-1}$ is the basis of $\Reals^d$, then $\bar\lambda_W$ is the embedding representation for $W$.
\end{restatable}

We can now connect to the intervention notion:
\begin{restatable}[Intervention Representation]{theorem}{intervention}
    \label{thm:intervention}
    Let $\bar{\lambda}_W$ be the embedding representation of a concept $W$. 
    Then, for any concept $Z$ that is causally separable with $W$,
    \begin{align*}
        \Pr(Y = Y(W, 1) \given Y\in\{Y(W, 0),Y(W, 1)\}, \lambda + c \bar{\lambda}_W)
    \end{align*}
    is constant in $c \in \Reals$, and
    \begin{align*}
        &\Pr(Y = Y(1, Z) \given Y\in\{Y(0, Z),Y(1, Z)\}, \lambda + c \bar{\lambda}_W)
    \end{align*}
    is increasing in $c \in \Reals$.
\end{restatable}
In words: adding $\bar{\lambda}_W$ to the language model representation of the context changes the probability of the target concept ($W$), but not the probability of off-target concepts ($Z$).

\section{Inner Product for Language Model Representations}\label{sec:inner_product}
Given linear representations, we would like to make use of them by doing things like measuring the similarity between different representations, or editing concepts by projecting onto a target direction.
Similarity and projection are both notions that require an inner product. 
We now consider the question of which inner product is appropriate for understanding language model representations.

\paragraph*{Preliminaries}
We define $\bar{\Gamma}$ to be the space of differences between elements of $\Gamma$.
Then, $\bar{\Gamma}$ is a $d$-dimensional real vector space.\footnote{Note that the unembedding space $\Gamma$ is only an affine space, since the softmax is invariant to adding a constant.}
We consider defining inner products on $\bar{\Gamma}$.
Unembedding representations are naturally directions (unique only up to scale).
Once we have an inner product, we define the \emph{canonical} unembedding representation $\bar{\gamma}_W$ to be the element of the cone with $\ip{\bar\gamma_W}{\bar\gamma_W} = 1$.
This lets us define inner products between unembedding representations.

\paragraph*{Unidentifiability of the inner product}
We might hope that there is some natural inner product that is picked out (identified) by the model training.
It turns out that this is not the case.
To understand the challenge, consider transforming the unembedding and embedding spaces according to
\begin{align}
    g(y) \leftarrow A \gamma(y) + \beta, \quad l(x) \leftarrow A^{-\top} \lambda(x),\label{eq:transform}
\end{align}
where $A \in \Reals^{d\times d}$ is some invertible linear transformation and $\beta \in \Reals^d$ is a constant.
It's easy to see that this transformation preserves the softmax distribution $\Pr(y \given x)$:
\begin{align*}
    \frac{\exp(\lambda(x)^\top \gamma(y))}{\sum_{y'}\exp(\lambda(x)^\top \gamma(y'))} = \frac{\exp(l(x)^\top g(y))}{\sum_{y'}\exp(l(x)^\top g(y'))}, \quad \forall x,y.
\end{align*}
However, the objective function used to train the model depends on the representations only through the softmax probabilities. 
Thus, the representation $\gamma$ is identified (at best) only up to some invertible affine transformation. 

This also means that the concept representations $\bar{\gamma}_W$ are identified only up to some invertible linear transformation $A$.
The problem is that, given any fixed inner product,
\begin{align*}
    \ip{\bar{\gamma}_W}{\bar{\gamma}_Z} \neq \ip{A\bar{\gamma}_W}{A\bar{\gamma}_Z},
\end{align*}
in general.
Accordingly, there is no obvious reason to expect that algebraic manipulations based on, e.g., the Euclidean inner product, should be semantically meaningful.

\subsection{Causal Inner Products}
We require some additional principles for choosing an inner product on the representation space.
The intuition we follow here is that causally separable concepts should be represented as orthogonal vectors.
For example, \ConceptDirName{English}{French} and \ConceptDirName{Male}{Female}, should be orthogonal.
We define an inner product with this property:
\begin{definition}
    [Causal Inner Product]\label{def:causal_inner_product}
    A \defnphrase{causal inner product} $\ip{\cdot}{\cdot}_{\mathrm{C}}$ on $\bar\Gamma \simeq \Reals^d$ is an inner product such that
    \begin{align*}
        \ip{\bar{\gamma}_W}{\bar{\gamma}_Z}_{\mathrm{C}} = 0,
    \end{align*}
    for any pair of causally separable concepts $W$ and $Z$.
\end{definition}

This choice turns out to have the key property that it unifies the unembedding and embedding representations:
\begin{restatable}[Unification of Representations]{theorem}{RepresentationUnification}\label{thm:isomorphism}
    Suppose that, for any concept $W$, there exist concepts $\{Z_i\}_{i=1}^{d-1}$ such that each $Z_i$ is causally separable with $W$ and $\{\bar\gamma_W\}\cup\{\bar\gamma_{Z_i}\}_{i=1}^{d-1}$ is a basis of $\Reals^d$. If $\ip{\cdot}{\cdot}_{\mathrm{C}}$ is a causal inner product, then the Riesz isomorphism $\bar\gamma \mapsto \ip{\bar\gamma}{\cdot}_{\mathrm{C}}$, for $\bar\gamma \in \bar\Gamma$, maps the unembedding representation $\bar\gamma_W$ of each concept $W$ to its embedding representation $\bar\lambda_W$:
    \begin{equation*}
        \ip{\bar{\gamma}_W}{\cdot}_{\mathrm{C}} = \bar\lambda_W^\top .
    \end{equation*}
\end{restatable}
To understand this result intuitively, notice we can represent embeddings as row vectors and unembeddings as column vectors. If the causal inner product were the Euclidean inner product, the isomorphism would simply be the transpose operation. The theorem is the (Riesz isomorphism) generalization of this idea: each linear map on $\bar{\Gamma}$ corresponds to some $\lambda \in \Lambda$ according to $\lambda^\top: \bar{\gamma} \mapsto \lambda^\top \bar{\gamma}$.
So, we can map $\bar{\Gamma}$ to $\Lambda$ by mapping each $\bar{\gamma}_W$ to a linear function according to $\bar{\gamma}_W \to \ip{\bar{\gamma}_W}{\cdot}_{\mathrm{C}}$. The theorem says this map sends each unembedding representation of a concept to the embedding representation of the same concept.

In the experiments, we will make use of this result to construct embedding representations from unembedding representations. In particular, this allows us to find interventional representations of concepts. This is important because it is difficult in practice to find pairs of prompts that directly satisfy \cref{def:embedding_rep}.

\subsection{An Explicit Form for Causal Inner Product}
The next problem is: if a causal inner product exists, how can we find it?
In principle, this could be done by finding the unembedding representations of a large number of concepts, and then finding an inner product that maps each pair of causally separable directions to zero.
In practice, this is infeasible because of the number of concepts required to find the inner product, and the difficulty of estimating the representations of each concept.

We now turn to developing a more tractable approach based on the following insight: knowing the value of concept $W$ expressed by a randomly chosen word tells us little about the value of a causally separable concept $Z$ expressed by that word.
For example, if we learn that a randomly sampled word is French (not English), this does not give us significant information about whether it refers to a man or woman.\footnote{Note that this assumption is about words \emph{sampled randomly from the vocabulary}, not words sampled randomly from natural language sources. In the latter, there may well be non-causal correlations between causally separable concepts.}
We formalize this idea as follows:
\begin{assumption}\label{ass:ortho_to_indep}
    Suppose $W,Z$ are causally separable concepts and that $\gamma$ is an unembedding vector sampled uniformly from the vocabulary. Then, $\bar{\lambda}_W^\top \gamma$ and $\bar{\lambda}_Z^\top \gamma$ are independent\footnote{In fact, to prove our next result, we only require that $\bar{\lambda}_W^\top \gamma$ and $\bar{\lambda}_Z^\top \gamma$ are uncorrelated. In Appendix~\ref{appendix:check_assumption}, we verify that the causal inner product we find satisfies the uncorrelatedness condition.} for any embedding representations $\bar{\lambda}_W$ and $\bar{\lambda}_Z$ for $W$ and $Z$, respectively.
\end{assumption}
This assumption lets us connect causal separability with something we can actually measure: the statistical dependency between words. The next result makes this precise.

\begin{restatable}[Explicit Form of Causal Inner Product]{theorem}{ExplicitCIP}\label{thm:explicit_form}
    Suppose there exists a causal inner product, represented as $\ip{\bar{\gamma}}{\bar{\gamma}'}_{\mathrm{C}} = \bar{\gamma}^\top M \bar{\gamma}'$ for some symmetric positive definite matrix $M$.
    If there are mutually causally separable concepts $\{W_k\}_{k=1}^d$, such that their canonical representations $G = [\bar{\gamma}_{W_1}, \cdots, \bar{\gamma}_{W_d}]$ form a basis for $\bar\Gamma\simeq \Reals^d$, then under \cref{ass:ortho_to_indep},
    \begin{equation}\label{eq:diagonalized}
        M^{-1} = G G^{\top} \text{ and } G^\top  \cov(\gamma)^{-1} G = D,
    \end{equation}
    for some diagonal matrix $D$ with positive entries, where $\gamma$ is the unembedding vector of a word sampled uniformly at random from the vocabulary.
\end{restatable}

Notice that causal orthogonality only imposes $d(d-1)/2$ constraints on the inner product, but there are $d(d-1)/2 + d$ degrees of freedom in identifying the positive definite matrix $M$ (hence, an inner product)---thus, we expect $d$ degrees of freedom in choosing a causal inner product. 
\Cref{thm:explicit_form} gives a characterization of this class of inner products, in the form of~\eqref{eq:diagonalized}. Here, $D$ is a free parameter with $d$ degrees of freedom. Each $D$ defines the inner product. We do not have a principle for picking out a unique choice of $D$.
In our experiments, we will work with the \emph{choice} $D = I_{d}$, which gives us $M = \cov(\gamma)^{-1}$.
Then, we have a simple closed form for the corresponding inner product: 
\begin{equation}\label{eq:our_CIP}
    \ip{\bar\gamma}{\bar\gamma'}_{\mathrm{C}} := \bar\gamma^\top \cov(\gamma)^{-1} \bar\gamma', \quad \forall \bar\gamma, \bar\gamma' \in \bar\Gamma.
\end{equation}
Note that although we don't have a unique inner product, we can rule out most inner products. E.g., the Euclidean inner product is not a causal inner product if $M=I_{d}$ does not satisfy~\eqref{eq:diagonalized} for any $D$.

\paragraph*{Unified representations}
The choice of inner product can also be viewed as defining a choice of representations $g$ and $l$ in \cref{eq:transform} (hence, $\bar{g} = A\bar{\gamma}$).
With $A = M^{1/2}$, Theorem~\ref{thm:isomorphism} further implies that a \emph{causal} inner product makes the embedding and unembedding representations of concepts the same, that is, $\bar{g}_W = \bar{l}_W$. 
Moreover, in the transformed space, the Euclidean inner product \emph{is} the causal inner product: $\ip{\bar\gamma}{\bar\gamma'}_{\mathrm{C}} = \bar{g}^\top \bar{g}'$.
In \cref{fig:sphere}, we illustrated this unification of unembedding and embedding representations.
This is convenient for experiments, because it allows the use of standard Euclidean tools on the transformed space.

\section{Experiments}
We now turn to empirically validating the existence of linear representations, the estimated causal inner product, and the predicted relationships between the subspace, measurement, and intervention notions of linear representation.
Code is available at \href{https://github.com/KihoPark/linear_rep_geometry}{github.com/KihoPark/linear\_rep\_geometry}.

We use the LLaMA-2 model with 7 billion parameters \citep{touvron2023llama2} as our testbed.
This is a decoder-only Transformer LLM \citep{vaswani2017attention,radford2018improving}, 
trained using the forward LM objective and a 32K token vocabulary.
We include further details on all experiments in \Cref{sec:exp_details}.

\paragraph{Concepts are represented as directions in the unembedding space}\label{sec:exp_subspace}
We start with the hypothesis that concepts are represented as directions in the unembedding representation space (\cref{def:unembedding_rep}).
This notion relies on counterfactual pairs of words that vary only in the value of the concept of interest.
We consider 22 concepts defined in the Big Analogy Test Set (BATS 3.0)~\citep{gladkova2016analogy}, which provides such counterfactual pairs.\footnote{We only utilize words that are single tokens in the LLaMA-2 model. See Appendix~\ref{sec:exp_details} for details.}
We also consider 4 language concepts: \ConceptDirName{English}{French}, \ConceptDirName{French}{German}, \ConceptDirName{French}{Spanish}, and \ConceptDirName{German}{Spanish}, where we use words and their translations as counterfactual pairs. 
Additionally, we consider the concept \ConceptDirName{frequent}{infrequent} capturing how common a word is---we use pairs of common/uncommon synonyms (e.g., ``bad'' and ``terrible'') as counterfactual pairs.
We provide a table of all 27 concepts we consider in \Cref{sec:exp_details}.

If the subspace notion of the linear representation hypothesis holds, then all counterfactual token pairs should point to a common direction in the unembedding space.
In practice, this will only hold approximately. However, if the linear representation hypothesis holds, we still expect that, e.g., $\gamma(\text{``queen''}) - \gamma(\text{``king''})$ will align with the \ConceptDirName{male}{female} direction (more closely than the difference between random word pairs will). 
To validate this, for each concept $W$, we look at how the direction defined by each counterfactual pair, $\gamma(y_i(1)) - \gamma(y_i(0))$, is geometrically aligned with the unembedding representation $\bar{\gamma}_W$.
We estimate $\bar{\gamma}_W$ as the (normalized) mean\footnote{Previous work on word embeddings~\citep{drozd2016word,fournier2020analogies} motivate taking the mean to improve the consistency of the concept direction.} among all counterfactual pairs: $\bar{\gamma}_W := \tilde{\gamma}_W/\sqrt{\ip{\tilde{\gamma}_W}{\tilde{\gamma}_W}_{\mathrm{C}}}$, where
\begin{equation*}
\tilde{\gamma}_W = \frac{1}{n_W} \sum_{i=1}^{n_W} \left[ \gamma(y_i(1)) - \gamma(y_i(0)) \right],
\end{equation*}
$n_W$ denotes the number of counterfactual pairs for $W$, and $\ip{\cdot}{\cdot}_\mathrm{C}$ denotes the causal inner product defined in~\eqref{eq:our_CIP}.

\begin{figure}[t]
    \centering
    \includegraphics[width=0.99\linewidth]{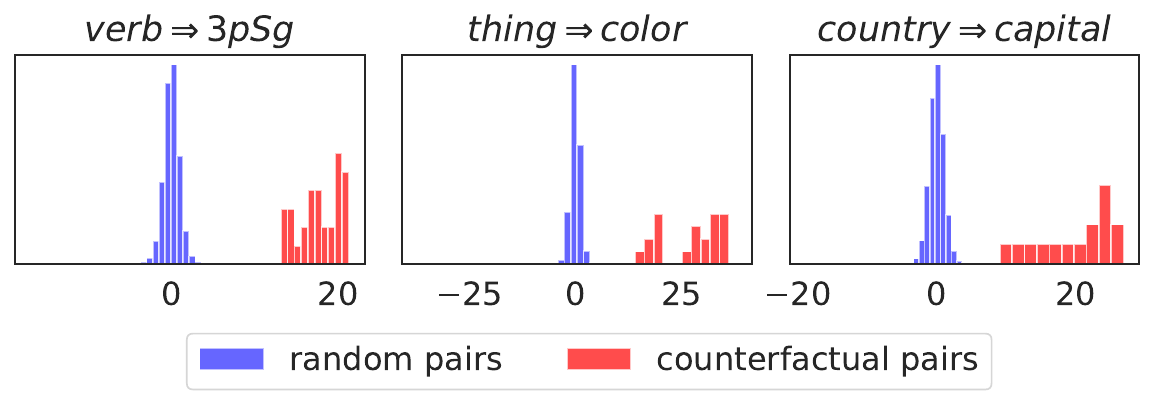}
    \caption{Projecting counterfactual pairs onto their corresponding concept direction shows a strong right skew, as we expect if the linear representation hypothesis holds. The projections of the counterfactual pairs, $\ip{\bar{\gamma}_{W, (-i)}}{\gamma(y_i(1)) - \gamma(y_i(0))}_{\mathrm{C}}$, are shown in red.
    For reference, we also project the differences between 100K randomly sampled word pairs onto the estimated concept direction, as shown in blue.
    See \Cref{tbl:concept_names} for details about each concept $W$ (the title of each plot).
    }
    \label{fig:right-skewed}
\end{figure}

\Cref{fig:right-skewed} presents histograms of each $\gamma(y_i(1)) - \gamma(y_i(0)))$ projected onto $\bar{\gamma}_W$ with respect to the causal inner product.
Since $\bar{\gamma}_W$ is computed using $\gamma(y_i(1)) - \gamma(y_i(0))$, we compute each projection using a leave-one-out (LOO) estimate $\bar{\gamma}_{W, (-i)}$ of the concept direction that excludes $(y_i(0), y_i(1))$.
Across the three concepts shown (and 23 others shown in \Cref{sec:additional_histograms}), the differences between counterfactual pairs are substantially more aligned with $\bar{\gamma}_W$ than those between random pairs.
The sole exception is \ConceptDirName{thing}{part}, which does not appear to have a linear representation.

The results are consistent with the linear representation hypothesis: the differences computed by each counterfactual pair point to a common direction representing a linear subspace (up to some noise). Further, $\bar{\gamma}_W$ is a reasonable estimator for that direction.

\paragraph{The estimated inner product respects causal separability}\label{sec:exp_ortho}

\begin{figure}[t]
    \centering
    \includegraphics[width=0.99\linewidth]{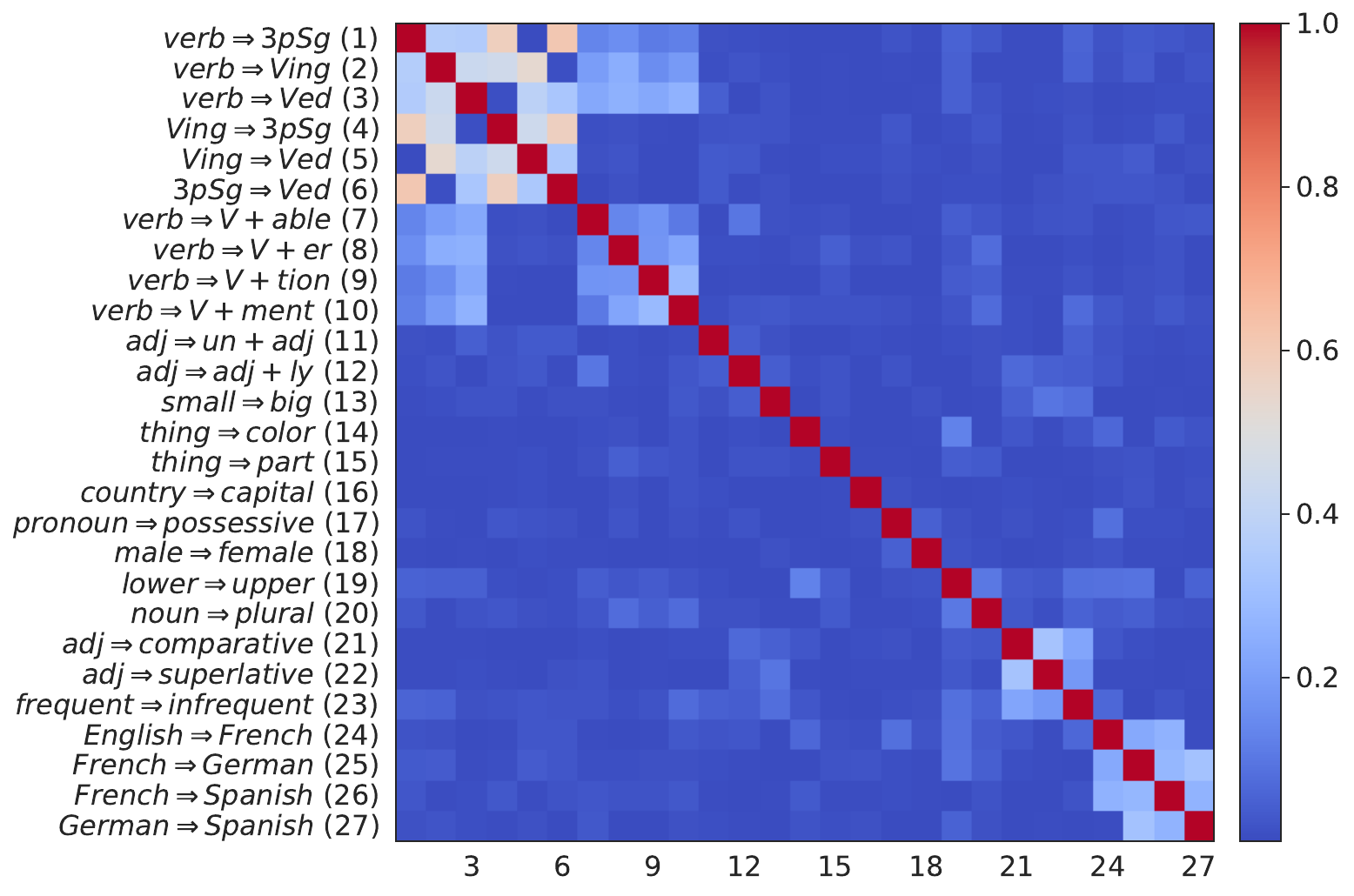}
    \caption{Causally separable concepts are represented approximately orthogonally under the estimated causal inner product based on~\eqref{eq:our_CIP}.
    The heatmap shows $ |\ip{\bar\gamma_W}{\bar\gamma_Z}_{\mathrm{C}}|$ for the estimated unembedding representations of each concept pair $(W, Z)$.
    The detail for each concept is given in \Cref{tbl:concept_names}.}
    \label{fig:heatmap_g_whitened}
\end{figure}

Next, we directly examine whether the estimated inner product~\eqref{eq:our_CIP} chosen from \cref{thm:explicit_form} is indeed approximately a causal inner product.
In \cref{fig:heatmap_g_whitened}, we plot a heatmap of the inner products between all pairs of the estimated unembedding representations for the 27 concepts.
If the estimated inner product is a causal inner product, then we expect values near 0 between causally separable concepts.

The first observation is that most pairs of concepts are nearly orthogonal with respect to this inner product. Interestingly, there is also a clear block diagonal structure. This arises because the concepts are grouped by semantic similarity. For example, the first 10 concepts relate to verbs, and the last 4 concepts are language pairs. The additional non-zero structure also generally makes sense. For example, \ConceptDirName{lower}{upper} (capitalization, concept 19) has non-trivial inner product with the language pairs \emph{other than} \ConceptDirName{French}{Spanish}. This may be because French and Spanish obey similar capitalization rules, while English and German each have different conventions (e.g., German capitalizes all nouns, but English only capitalizes proper nouns).
In \Cref{appendix:other_inner_product}, we compare the Euclidean inner product to the causal inner product for both the LLaMA-2 model and a more recent Gemma large language model \Citep{team2024gemma}.

\begin{figure}[t]
    \centering
    \includegraphics[width=0.99\linewidth]{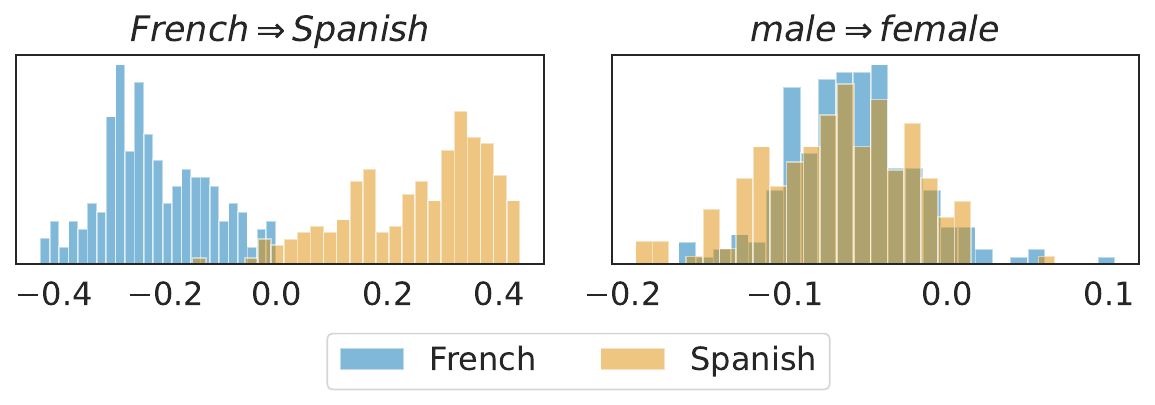}
    \caption{The subspace representation $\bar\gamma_W$ acts as a linear probe for $W$. The histograms show $\bar\gamma_W^\top \lambda(x_j^\texttt{fr})$ vs.~$\bar\gamma_W^\top \lambda(x_j^\texttt{es})$ (left) and $\bar\gamma_Z^\top \lambda(x_j^\texttt{fr})$ vs.~$\bar\gamma_Z^\top \lambda(x_j^\texttt{es})$ (right) for $W=$ \ConceptDirName{French}{Spanish} and $Z=$ \ConceptDirName{male}{female}, where $\{x_j^\texttt{fr}\}$ and $\{x_j^\texttt{es}\}$ are random contexts from French and Spanish Wikipedia, respectively.
    We also see that $\bar\gamma_Z$ does \emph{not} act as a linear probe for $W$, as expected.
    }\label{fig:measurement}
\end{figure}

\paragraph{Concept directions act as linear probes}\label{sec:exp_measurement}
Next, we check the connection to the measurement notion of linear representation. 
We consider the concept $W=$ \ConceptDirName{French}{Spanish}. 
To construct a dataset of French and Spanish contexts, we sample contexts
of random lengths from Wikipedia pages in each language.
Note that these are \emph{not} counterfactual pairs.
Following \cref{thm:measurement}, we expect $\bar{\gamma}_W^\top \lambda(x_j^\texttt{fr}) <0$ and $\bar{\gamma}_W^\top \lambda(x_j^\texttt{es})>0$. \Cref{fig:measurement} confirms this expectation, showing that $\bar\gamma_W$ is a linear probe for the concept $W$ in $\Lambda$ (left).
Also, the representation of an off-target concept $Z=$ \ConceptDirName{male}{female} does not have any predictive power for this task (right). 
\Cref{sec:additional_measurements} includes analogous results using all 27 concepts.

\paragraph{Concept directions map to intervention representations}\label{sec:exp_intervention}
\Cref{thm:intervention} says that we can construct an intervention representation by constructing an embedding representation.
Doing this directly requires finding pairs of prompts that vary only on the distribution they induce on the target concept, which can be difficult to find in practice. 

Here, we will instead use the isomorphism between embedding and unembedding representations (\cref{thm:isomorphism}) to construct intervention representations from unembedding representations.
We take
\begin{equation}\label{eq:new_intervention_rep}
    \bar\lambda_W : = \cov(\gamma)^{-1}\bar\gamma_W.
\end{equation}
\Cref{thm:intervention} predicts that adding $\bar\lambda_W$ to a context representation should increase the probability of $W$, while leaving the probability of all causally separable concepts unaltered. 

To test this for a given pair of causally separable concepts $W$ and $Z$, we first choose a quadruple $\{Y(w,z)\}_{w,z\in\{0,1\}}$, and then generate contexts $\{x_j\}$ such that the next word should be $Y(0,0)$. 
For example, if $W=$ \ConceptDirName{male}{female} and $Z=$ \ConceptDirName{lower}{upper}, then we choose the quadruple (``king'', ``queen'', ``King'', ``Queen''), and generate contexts using ChatGPT-4 (e.g., ``Long live the'').
We then intervene on $\lambda(x_j)$ using $\bar\lambda_{C}$ via
\begin{equation}\label{eqn:intervention}
    \lambda_{C,\alpha}(x_j) = \lambda(x_j) + \alpha \bar\lambda_{C},
\end{equation}
where $\alpha > 0$ and $C$ can be $W$, $Z$, or some other causally separable concept (e.g., \ConceptDirName{French}{Spanish}).
For different choices of $C$, we plot the changes in $\logit \Pr(W=1 \given Z, \lambda)$ and $\logit \Pr(Z=1 \given W, \lambda)$, as we increase $\alpha$.
We expect to see that, if we intervene in the $W$ direction, then the intervention should linearly increase $\logit \Pr(W=1 \given Z, \lambda)$, while the other logit should stay constant; if we intervene in a direction $C$ that is causally separable with both $W$ and $Z$, then we expect both logits to stay constant.

\begin{figure}[t]
    \centering
    \includegraphics[width=0.99\linewidth]{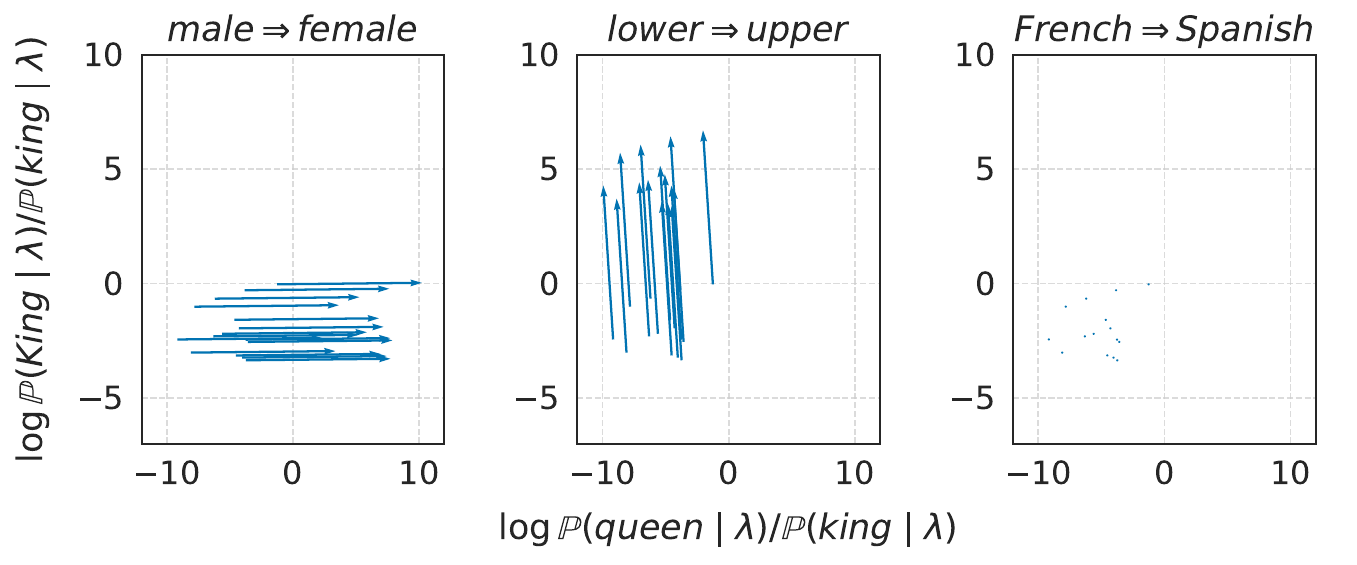}
    \caption{Adding $\alpha\bar\lambda_C$ to $\lambda$ changes the target concept $C$ without changing off-target concepts. The plots illustrate change in $\log(\Pr(\mathrm{``queen"}  \mid x ) / \Pr(\mathrm{``king"}  \mid x ))$ and $\log(\Pr(\mathrm{``King"}  \mid x ) / \Pr(\mathrm{``king"}  \mid x ))$, after changing $\lambda(x_j)$ to $\lambda_{C,\alpha}(x_j)$ as $\alpha$ increases from $0$ to $0.4$, for $C = $ \ConceptDirName{male}{female} (left), \ConceptDirName{lower}{upper} (center), \ConceptDirName{French}{Spanish} (right). The two ends of the arrow are $\lambda(x_j)$ and $\lambda_{C,0.4}(x_j)$, respectively. Each context $x_j$ is presented in \Cref{tbl:intervention_texts}.
    }
    \label{fig:intervention_flip}
\end{figure}

\Cref{fig:intervention_flip} shows the results of one such experiment shown for three target concepts (24 others shown in \Cref{sec:additional_intervention}), confirming our expectations. 
We see, for example, that intervening in the \ConceptDirName{male}{female} direction raises the logit for choosing ``queen'' over ``king'' as the next word, but does not change the logit for ``King'' over ``king''.

\begin{table}[t]
\caption{Adding the intervention representation $\alpha \bar\lambda_W$ pushes the probability over completions to reflect the concept $W$.
    As the scale of intervention increases, the probability of seeing $Y(W=1)$ (``\textbf{queen}'') increases while the probability of seeing $Y(W=0)$ (``\textit{king}'') decreases. We show the top-5 most probable words over the entire vocabulary following the intervention~\eqref{eqn:intervention} in the $W=$ \ConceptDirName{male}{female} direction, i.e., $\lambda_{W,\alpha}(x) = \lambda(x) + \alpha \bar\lambda_W$, for $\alpha \in \{0, 0.1, 0.2, 0.3, 0.4\}$.
    The original context $x=$ ``Long live the '' is a sentence fragment that ends with the word $Y(W=0)$ (``\textit{king}'').
    The most likely words reflect the concept, with ``\textbf{queen}'' being top-1. In \Cref{sec:additional_tables}, we provide more examples.}
\label{tbl:intervention_topk}
\vskip 0.15in
\begin{center}
\begin{small}
\begin{tabular}{cccccc}
\toprule
Rank  & $\alpha =$ 0 & 0.1 & 0.2 & 0.3 & 0.4\\
\midrule
1 & \it king & Queen & \bf queen & \bf queen & \bf queen \\
2 & King & \bf queen & Queen & Queen & Queen \\
3 & Queen & \it king & \_ & lady & lady \\
4 & \bf queen & King & lady & woman & woman \\
5 & \_ & \_ & \it king & women & women \\
\bottomrule
\end{tabular}
\end{small}
\end{center}
\vskip -0.1in
\end{table}

A natural follow-up question is to see if the intervention in a concept direction (for $W$) pushes the probability of $Y(W=1)$ being the next word to be the largest among all tokens.
We expect to see that, as we increase the value of $\alpha$, the target concept should eventually be reflected in the most likely output words according to the LM.

In \Cref{tbl:intervention_topk}, we show an illustrative example in which $W$ is the concept \ConceptDirName{male}{female} and the context $x$ is a sentence fragment that can end with the word $Y(W=0)$ (``\textit{king}'').
For $x =$ ``Long live the '', as we increase the scale $\alpha$ on the intervention, we see that the target word $Y(W=1)$ (``\textbf{queen}'') becomes the most likely next word, while the original word $Y(W=0)$ drops below the top-5 list.
This illustrates how the intervention can push the probability of the target word high enough to make it the most likely word while decreasing the probability of the original word.

\section{Discussion and Related Work}\label{sec:relatedwork}
The idea that high-level concepts are encoded \emph{linearly} is appealing because---if it is true---it may open up simple methods for interpretation and control of LLMs.
In this paper, we have formalized `linear representation', and shown that all natural variants of this notion can be unified.\footnote{In \Cref{sec:summary}, we summarize these results in a figure.} 
This equivalence already suggests some approaches for interpretation and control---e.g., we show how to use collections of pairs of words to define concept directions,
and then use these directions to predict what the model's output will be,
and to change the output in a controlled fashion.
A major theme is the role played by the choice of inner product.

\paragraph{Linear subspaces in language representations}
The linear subspace hypotheses was originally observed empirically in the context of word embeddings \citep[e.g.,][]{mikolov2013distributed,mikolov2013linguistic,levy2014linguistic,goldberg2014word2vec,vylomova2016take,gladkova2016analogy,chiang2020understanding, fournier2020analogies}. Similar structure has been observed in cross-lingual word embeddings~\citep{mikolov2013exploiting,lample2018word,ruder2019survey,peng2022understanding}, sentence embeddings~\citep{bowman2016generating,zhu2020sentence,li2020sentence,ushio2021bert}, representation spaces of Transformer LLMs~\citep{meng2022locating,merullo2023language,hernandez2023linearity}, and vision-language models~\citep{wang2023concept,trager2023linear,perera2023prompt}.
These observations motivate \Cref{def:unembedding_rep}. 
The key idea in the present paper is providing formalization in terms of counterfactual pairs---this is what allows us to connect to other notions of linear representation, and to identify the inner product structure.

\paragraph{Measurement, intervention, and mechanistic interpretability}
There is a significant body of work on linear representations for interpreting (probing) \citep[e.g.,][]{alain2017understanding, kim2018interpretability,nostalgebraist2020logitlens,rogers2021primer, belinkov2022probing, li2022emergent, geva2022transformer, nanda2023emergent}
and controlling (steering) \citep[e.g.,][]{wang2023concept,ActivationAddition:2023,merullo2023language,trager2023linear} models. This is particularly prominent in \emph{mechanistic interpretability} \citep{elhage2021mathematical,meng2022locating,hernandez2023linearity,ActivationAddition:2023,zou2023representation,todd2023function,hendel2023incontext}. 
With respect to this body of work, the main contribution of the present paper is to clarify the linear representation hypothesis, and the critical role of the inner product. 
However, we do not address interpretability of either model parameters, nor the activations of intermediate layers. These are main focuses of existing work. It is an exciting direction for future work to understand how ideas here---particularly, the causal inner product---translate to these settings.

\paragraph{Geometry of representations}
There is a line of work that studies the geometry of word and sentence representations \citep[e.g.,][]{arora2016latent,mimno2017strange,ethayarajh2019contextual,reif2019visualizing,li2020sentence,hewitt2019structural,chen2021probing, chang2022geometry, jiang2023uncovering}.
This work considers, e.g., visualizing and modeling how the learned embeddings are distributed, or how hierarchical structure is encoded.
Our work is largely orthogonal to these, since we are attempting to define a suitable inner product (and thus, notions of similarity and projection) that respects the semantic structure of language.

\paragraph{Causal representation learning}
Finally, the ideas here connect to causal representation learning~\citep[e.g.,][]{higgins2016beta,hyvarinen2016unsupervised,higgins2018towards,khemakhem2020variational,zimmermann2021contrastive,scholkopf2021toward,MoranDeepGenIdent:2021,wang2023concept}.
Most obviously, our causal formalization of concepts is inspired by \citet{wang2023concept}, who establish a characterization of latent concepts and vector algebra in diffusion models.
Separately, a major theme in this literature is the identifiability of learned representations---i.e., to what extent they capture underlying real-world structure.
Our causal inner product results may be viewed in this theme, showing that an inner product respecting semantic closeness is not identified by the usual training procedure, but that it can be picked out with a suitable assumption.  

\section*{Acknowledgements}
Thanks to Gemma Moran for comments on an earlier draft.
This work is supported by ONR grant N00014-23-1-2591 and Open Philanthropy.

\bibliography{bib/concept}

\begin{thebibliography}{61}
\providecommand{\natexlab}[1]{#1}
\providecommand{\url}[1]{\texttt{#1}}
\expandafter\ifx\csname urlstyle\endcsname\relax
  \providecommand{\doi}[1]{doi: #1}\else
  \providecommand{\doi}{doi: \begingroup \urlstyle{rm}\Url}\fi

\bibitem[Alain \& Bengio(2017)Alain and Bengio]{alain2017understanding}
Alain, G. and Bengio, Y.
\newblock Understanding intermediate layers using linear classifier probes.
\newblock In \emph{International Conference on Learning Representations}, 2017.
\newblock URL \url{https://openreview.net/forum?id=ryF7rTqgl}.

\bibitem[Arora et~al.(2016)Arora, Li, Liang, Ma, and Risteski]{arora2016latent}
Arora, S., Li, Y., Liang, Y., Ma, T., and Risteski, A.
\newblock A latent variable model approach to {PMI}-based word embeddings.
\newblock \emph{Transactions of the Association for Computational Linguistics},
  4:\penalty0 385--399, 2016.

\bibitem[Belinkov(2022)]{belinkov2022probing}
Belinkov, Y.
\newblock Probing classifiers: Promises, shortcomings, and advances.
\newblock \emph{Computational Linguistics}, 48\penalty0 (1):\penalty0 207--219,
  2022.

\bibitem[Bowman et~al.(2016)Bowman, Vilnis, Vinyals, Dai, Jozefowicz, and
  Bengio]{bowman2016generating}
Bowman, S.~R., Vilnis, L., Vinyals, O., Dai, A., Jozefowicz, R., and Bengio, S.
\newblock Generating sentences from a continuous space.
\newblock In \emph{Proceedings of the 20th {SIGNLL} Conference on Computational
  Natural Language Learning}, pp.\  10--21, Berlin, Germany, August 2016.
  Association for Computational Linguistics.
\newblock \doi{10.18653/v1/K16-1002}.
\newblock URL \url{https://aclanthology.org/K16-1002}.

\bibitem[Chang et~al.(2022)Chang, Tu, and Bergen]{chang2022geometry}
Chang, T., Tu, Z., and Bergen, B.
\newblock The geometry of multilingual language model representations.
\newblock In \emph{Proceedings of the 2022 Conference on Empirical Methods in
  Natural Language Processing}, pp.\  119--136, 2022.

\bibitem[Chen et~al.(2021)Chen, Fu, Xu, Xie, Tan, Chen, and
  Jing]{chen2021probing}
Chen, B., Fu, Y., Xu, G., Xie, P., Tan, C., Chen, M., and Jing, L.
\newblock Probing {BERT} in hyperbolic spaces.
\newblock In \emph{International Conference on Learning Representations}, 2021.

\bibitem[Chiang et~al.(2020)Chiang, Camacho-Collados, and
  Pardos]{chiang2020understanding}
Chiang, H.-Y., Camacho-Collados, J., and Pardos, Z.
\newblock Understanding the source of semantic regularities in word embeddings.
\newblock In \emph{Proceedings of the 24th Conference on Computational Natural
  Language Learning}, pp.\  119--131, 2020.

\bibitem[Choe et~al.(2020)Choe, Park, and Kim]{choe2020word2word}
Choe, Y.~J., Park, K., and Kim, D.
\newblock word2word: A collection of bilingual lexicons for 3,564 language
  pairs.
\newblock In \emph{Proceedings of the Twelfth Language Resources and Evaluation
  Conference}, pp.\  3036--3045, 2020.

\bibitem[Drozd et~al.(2016)Drozd, Gladkova, and Matsuoka]{drozd2016word}
Drozd, A., Gladkova, A., and Matsuoka, S.
\newblock Word embeddings, analogies, and machine learning: Beyond {king - man
  + woman = queen}.
\newblock In \emph{Proceedings of COLING 2016, the 26th International
  Conference on Computational Linguistics: Technical papers}, pp.\  3519--3530,
  2016.

\bibitem[Elhage et~al.(2021)Elhage, Nanda, Olsson, Henighan, Joseph, Mann,
  Askell, Bai, Chen, Conerly, et~al.]{elhage2021mathematical}
Elhage, N., Nanda, N., Olsson, C., Henighan, T., Joseph, N., Mann, B., Askell,
  A., Bai, Y., Chen, A., Conerly, T., et~al.
\newblock A mathematical framework for transformer circuits.
\newblock \emph{Transformer Circuits Thread}, 1, 2021.

\bibitem[Elhage et~al.(2022)Elhage, Hume, Olsson, Schiefer, Henighan, Kravec,
  Hatfield-Dodds, Lasenby, Drain, Chen, et~al.]{elhage2022toy}
Elhage, N., Hume, T., Olsson, C., Schiefer, N., Henighan, T., Kravec, S.,
  Hatfield-Dodds, Z., Lasenby, R., Drain, D., Chen, C., et~al.
\newblock Toy models of superposition.
\newblock \emph{arXiv preprint arXiv:2209.10652}, 2022.

\bibitem[Ethayarajh(2019)]{ethayarajh2019contextual}
Ethayarajh, K.
\newblock How contextual are contextualized word representations? {Comparing}
  the geometry of {BERT}, {ELMo}, and {GPT-2} embeddings.
\newblock In \emph{Proceedings of the 2019 Conference on Empirical Methods in
  Natural Language Processing and the 9th International Joint Conference on
  Natural Language Processing (EMNLP-IJCNLP)}, pp.\  55--65, 2019.

\bibitem[Fournier et~al.(2020)Fournier, Dupoux, and
  Dunbar]{fournier2020analogies}
Fournier, L., Dupoux, E., and Dunbar, E.
\newblock Analogies minus analogy test: measuring regularities in word
  embeddings.
\newblock In \emph{Proceedings of the 24th Conference on Computational Natural
  Language Learning}, pp.\  365--375, Online, 2020. Association for
  Computational Linguistics.
\newblock \doi{10.18653/v1/2020.conll-1.29}.
\newblock URL \url{https://aclanthology.org/2020.conll-1.29}.

\bibitem[Geva et~al.(2022)Geva, Caciularu, Wang, and
  Goldberg]{geva2022transformer}
Geva, M., Caciularu, A., Wang, K., and Goldberg, Y.
\newblock Transformer feed-forward layers build predictions by promoting
  concepts in the vocabulary space.
\newblock In \emph{Proceedings of the Conference on Empirical Methods in
  Natural Language Processing}, pp.\  30--45, 2022.

\bibitem[Gladkova et~al.(2016)Gladkova, Drozd, and
  Matsuoka]{gladkova2016analogy}
Gladkova, A., Drozd, A., and Matsuoka, S.
\newblock Analogy-based detection of morphological and semantic relations with
  word embeddings: what works and what doesn’t.
\newblock In \emph{Proceedings of the NAACL Student Research Workshop}, pp.\
  8--15, 2016.

\bibitem[Goldberg \& Levy(2014)Goldberg and Levy]{goldberg2014word2vec}
Goldberg, Y. and Levy, O.
\newblock word2vec explained: deriving {Mikolov} et al.'s negative-sampling
  word-embedding method.
\newblock \emph{arXiv preprint arXiv:1402.3722}, 2014.

\bibitem[Gurnee \& Tegmark(2023)Gurnee and Tegmark]{LMSpaceTime:2023}
Gurnee, W. and Tegmark, M.
\newblock Language models represent space and time.
\newblock \emph{arXiv preprint arXiv:2310.02207}, art. arXiv:2310.02207,
  October 2023.
\newblock \doi{10.48550/arXiv.2310.02207}.

\bibitem[Hendel et~al.(2023)Hendel, Geva, and Globerson]{hendel2023incontext}
Hendel, R., Geva, M., and Globerson, A.
\newblock In-context learning creates task vectors.
\newblock \emph{arXiv preprint arXiv:2310.15916}, 2023.

\bibitem[Hernandez et~al.(2023)Hernandez, Sharma, Haklay, Meng, Wattenberg,
  Andreas, Belinkov, and Bau]{hernandez2023linearity}
Hernandez, E., Sharma, A.~S., Haklay, T., Meng, K., Wattenberg, M., Andreas,
  J., Belinkov, Y., and Bau, D.
\newblock Linearity of relation decoding in transformer language models.
\newblock \emph{arXiv preprint arXiv:2308.09124}, 2023.

\bibitem[Hewitt \& Manning(2019)Hewitt and Manning]{hewitt2019structural}
Hewitt, J. and Manning, C.~D.
\newblock A structural probe for finding syntax in word representations.
\newblock In \emph{Proceedings of the 2019 Conference of the North American
  Chapter of the Association for Computational Linguistics: Human Language
  Technologies, Volume 1 (Long and Short Papers)}, pp.\  4129--4138, 2019.

\bibitem[Higgins et~al.(2016)Higgins, Matthey, Pal, Burgess, Glorot, Botvinick,
  Mohamed, and Lerchner]{higgins2016beta}
Higgins, I., Matthey, L., Pal, A., Burgess, C., Glorot, X., Botvinick, M.,
  Mohamed, S., and Lerchner, A.
\newblock beta-{VAE}: Learning basic visual concepts with a constrained
  variational framework.
\newblock In \emph{International Conference on Learning Representations}, 2016.

\bibitem[Higgins et~al.(2018)Higgins, Amos, Pfau, Racaniere, Matthey, Rezende,
  and Lerchner]{higgins2018towards}
Higgins, I., Amos, D., Pfau, D., Racaniere, S., Matthey, L., Rezende, D., and
  Lerchner, A.
\newblock Towards a definition of disentangled representations.
\newblock \emph{arXiv preprint arXiv:1812.02230}, 2018.

\bibitem[Hyvarinen \& Morioka(2016)Hyvarinen and
  Morioka]{hyvarinen2016unsupervised}
Hyvarinen, A. and Morioka, H.
\newblock Unsupervised feature extraction by time-contrastive learning and
  nonlinear {ICA}.
\newblock \emph{Advances in Neural Information Processing Systems}, 29, 2016.

\bibitem[Jiang et~al.(2023)Jiang, Aragam, and Veitch]{jiang2023uncovering}
Jiang, Y., Aragam, B., and Veitch, V.
\newblock Uncovering meanings of embeddings via partial orthogonality.
\newblock \emph{arXiv preprint arXiv:2310.17611}, 2023.

\bibitem[Khemakhem et~al.(2020)Khemakhem, Kingma, Monti, and
  Hyvarinen]{khemakhem2020variational}
Khemakhem, I., Kingma, D., Monti, R., and Hyvarinen, A.
\newblock Variational autoencoders and nonlinear {ICA}: A unifying framework.
\newblock In \emph{International Conference on Artificial Intelligence and
  Statistics}, pp.\  2207--2217. PMLR, 2020.

\bibitem[Kim et~al.(2018)Kim, Wattenberg, Gilmer, Cai, Wexler, Viegas,
  et~al.]{kim2018interpretability}
Kim, B., Wattenberg, M., Gilmer, J., Cai, C., Wexler, J., Viegas, F., et~al.
\newblock Interpretability beyond feature attribution: Quantitative testing
  with concept activation vectors ({TCAV}).
\newblock In \emph{International Conference on Machine Learning}, pp.\
  2668--2677. PMLR, 2018.

\bibitem[Kudo \& Richardson(2018)Kudo and Richardson]{kudo2018sentencepiece}
Kudo, T. and Richardson, J.
\newblock {SentencePiece}: A simple and language independent subword tokenizer
  and detokenizer for neural text processing.
\newblock In \emph{Proceedings of the 2018 Conference on Empirical Methods in
  Natural Language Processing: System Demonstrations}, pp.\  66--71, 2018.

\bibitem[Lample et~al.(2018)Lample, Conneau, Ranzato, Denoyer, and
  J{\'e}gou]{lample2018word}
Lample, G., Conneau, A., Ranzato, M., Denoyer, L., and J{\'e}gou, H.
\newblock Word translation without parallel data.
\newblock In \emph{International Conference on Learning Representations}, 2018.

\bibitem[Levy \& Goldberg(2014)Levy and Goldberg]{levy2014linguistic}
Levy, O. and Goldberg, Y.
\newblock Linguistic regularities in sparse and explicit word representations.
\newblock In \emph{Proceedings of the Eighteenth Conference on Computational
  Natural Language Learning}, pp.\  171--180, 2014.

\bibitem[Li et~al.(2020)Li, Zhou, He, Wang, Yang, and Li]{li2020sentence}
Li, B., Zhou, H., He, J., Wang, M., Yang, Y., and Li, L.
\newblock On the sentence embeddings from pre-trained language models.
\newblock In \emph{Proceedings of the 2020 Conference on Empirical Methods in
  Natural Language Processing (EMNLP)}, pp.\  9119--9130, 2020.

\bibitem[Li et~al.(2022)Li, Hopkins, Bau, Vi{\'e}gas, Pfister, and
  Wattenberg]{li2022emergent}
Li, K., Hopkins, A.~K., Bau, D., Vi{\'e}gas, F., Pfister, H., and Wattenberg,
  M.
\newblock Emergent world representations: Exploring a sequence model trained on
  a synthetic task.
\newblock In \emph{International Conference on Learning Representations}, 2022.

\bibitem[Meng et~al.(2022)Meng, Bau, Andonian, and Belinkov]{meng2022locating}
Meng, K., Bau, D., Andonian, A., and Belinkov, Y.
\newblock Locating and editing factual associations in {GPT}.
\newblock \emph{Advances in Neural Information Processing Systems},
  35:\penalty0 17359--17372, 2022.

\bibitem[Merullo et~al.(2023)Merullo, Eickhoff, and
  Pavlick]{merullo2023language}
Merullo, J., Eickhoff, C., and Pavlick, E.
\newblock Language models implement simple word2vec-style vector arithmetic.
\newblock \emph{arXiv preprint arXiv:2305.16130}, 2023.

\bibitem[Mesnard et~al.(2024)Mesnard, Hardin, Dadashi, Bhupatiraju, Pathak,
  Sifre, Rivi{\`e}re, Kale, Love, et~al.]{team2024gemma}
Mesnard, T., Hardin, C., Dadashi, R., Bhupatiraju, S., Pathak, S., Sifre, L.,
  Rivi{\`e}re, M., Kale, M.~S., Love, J., et~al.
\newblock Gemma: Open models based on gemini research and technology.
\newblock \emph{arXiv preprint arXiv:2403.08295}, 2024.

\bibitem[Mikolov et~al.(2013{\natexlab{a}})Mikolov, Le, and
  Sutskever]{mikolov2013exploiting}
Mikolov, T., Le, Q.~V., and Sutskever, I.
\newblock Exploiting similarities among languages for machine translation.
\newblock \emph{arXiv preprint arXiv:1309.4168}, 2013{\natexlab{a}}.

\bibitem[Mikolov et~al.(2013{\natexlab{b}})Mikolov, Sutskever, Chen, Corrado,
  and Dean]{mikolov2013distributed}
Mikolov, T., Sutskever, I., Chen, K., Corrado, G.~S., and Dean, J.
\newblock Distributed representations of words and phrases and their
  compositionality.
\newblock \emph{Advances in Neural Information Processing Systems}, 26,
  2013{\natexlab{b}}.

\bibitem[Mikolov et~al.(2013{\natexlab{c}})Mikolov, Yih, and
  Zweig]{mikolov2013linguistic}
Mikolov, T., Yih, W.-T., and Zweig, G.
\newblock Linguistic regularities in continuous space word representations.
\newblock In \emph{Proceedings of the 2013 Conference of the North American
  Chapter of the Association for Computational Linguistics: Human Language
  Technologies}, pp.\  746--751, 2013{\natexlab{c}}.

\bibitem[Mimno \& Thompson(2017)Mimno and Thompson]{mimno2017strange}
Mimno, D. and Thompson, L.
\newblock The strange geometry of skip-gram with negative sampling.
\newblock In Palmer, M., Hwa, R., and Riedel, S. (eds.), \emph{Proceedings of
  the 2017 Conference on Empirical Methods in Natural Language Processing},
  pp.\  2873--2878, Copenhagen, Denmark, 2017. Association for Computational
  Linguistics.
\newblock \doi{10.18653/v1/D17-1308}.
\newblock URL \url{https://aclanthology.org/D17-1308}.

\bibitem[{Moran} et~al.(2021){Moran}, {Sridhar}, {Wang}, and
  {Blei}]{MoranDeepGenIdent:2021}
{Moran}, G.~E., {Sridhar}, D., {Wang}, Y., and {Blei}, D.~M.
\newblock Identifiable deep generative models via sparse decoding.
\newblock \emph{arXiv preprint arXiv:2110.10804}, art. arXiv:2110.10804,
  October 2021.
\newblock \doi{10.48550/arXiv.2110.10804}.

\bibitem[Nanda et~al.(2023)Nanda, Lee, and Wattenberg]{nanda2023emergent}
Nanda, N., Lee, A., and Wattenberg, M.
\newblock Emergent linear representations in world models of self-supervised
  sequence models.
\newblock \emph{arXiv preprint arXiv:2309.00941}, 2023.

\bibitem[nostalgebraist(2020)]{nostalgebraist2020logitlens}
nostalgebraist.
\newblock {Interpreting {GPT}: the logit lens}, 2020.
\newblock URL
  \url{https://www.alignmentforum.org/posts/AcKRB8wDpdaN6v6ru/interpreting-gpt-the-logit-lens}.

\bibitem[OpenAI(2023)]{openai2023gpt4}
OpenAI.
\newblock {GPT-4} technical report.
\newblock \emph{arXiv preprint arXiv:2303.08774}, 2023.

\bibitem[Peng et~al.(2022)Peng, Stevenson, Lin, and Li]{peng2022understanding}
Peng, X., Stevenson, M., Lin, C., and Li, C.
\newblock Understanding linearity of cross-lingual word embedding mappings.
\newblock \emph{Transactions on Machine Learning Research}, 2022.
\newblock ISSN 2835-8856.
\newblock URL \url{https://openreview.net/forum?id=8HuyXvbvqX}.

\bibitem[Pennington et~al.(2014)Pennington, Socher, and
  Manning]{pennington2014glove}
Pennington, J., Socher, R., and Manning, C.~D.
\newblock {GloVe: Global vectors for word representation}.
\newblock In \emph{Proceedings of the 2014 Conference on Empirical Methods in
  Natural Language Processing (EMNLP)}, pp.\  1532--1543, 2014.

\bibitem[Perera et~al.(2023)Perera, Trager, Zancato, Achille, and
  Soatto]{perera2023prompt}
Perera, P., Trager, M., Zancato, L., Achille, A., and Soatto, S.
\newblock Prompt algebra for task composition.
\newblock \emph{arXiv preprint arXiv:2306.00310}, 2023.

\bibitem[Radford et~al.(2018)Radford, Narasimhan, Salimans, and
  Sutskever]{radford2018improving}
Radford, A., Narasimhan, K., Salimans, T., and Sutskever, I.
\newblock Improving language understanding by generative pre-training.
\newblock 2018.

\bibitem[Reif et~al.(2019)Reif, Yuan, Wattenberg, Viegas, Coenen, Pearce, and
  Kim]{reif2019visualizing}
Reif, E., Yuan, A., Wattenberg, M., Viegas, F.~B., Coenen, A., Pearce, A., and
  Kim, B.
\newblock Visualizing and measuring the geometry of {BERT}.
\newblock \emph{Advances in Neural Information Processing Systems}, 32, 2019.

\bibitem[Rogers et~al.(2021)Rogers, Kovaleva, and Rumshisky]{rogers2021primer}
Rogers, A., Kovaleva, O., and Rumshisky, A.
\newblock {A primer in BERTology: What we know about how BERT works}.
\newblock \emph{Transactions of the Association for Computational Linguistics},
  8:\penalty0 842--866, 2021.

\bibitem[Ruder et~al.(2019)Ruder, Vuli{\'c}, and S{\o}gaard]{ruder2019survey}
Ruder, S., Vuli{\'c}, I., and S{\o}gaard, A.
\newblock A survey of cross-lingual word embedding models.
\newblock \emph{Journal of Artificial Intelligence Research}, 65:\penalty0
  569--631, 2019.

\bibitem[Sch{\"o}lkopf et~al.(2021)Sch{\"o}lkopf, Locatello, Bauer, Ke,
  Kalchbrenner, Goyal, and Bengio]{scholkopf2021toward}
Sch{\"o}lkopf, B., Locatello, F., Bauer, S., Ke, N.~R., Kalchbrenner, N.,
  Goyal, A., and Bengio, Y.
\newblock Toward causal representation learning.
\newblock \emph{Proceedings of the IEEE}, 109\penalty0 (5):\penalty0 612--634,
  2021.

\bibitem[Todd et~al.(2023)Todd, Li, Sharma, Mueller, Wallace, and
  Bau]{todd2023function}
Todd, E., Li, M.~L., Sharma, A.~S., Mueller, A., Wallace, B.~C., and Bau, D.
\newblock Function vectors in large language models.
\newblock \emph{arXiv preprint arXiv:2310.15213}, 2023.

\bibitem[Touvron et~al.(2023)Touvron, Martin, Stone, Albert, Almahairi, Babaei,
  Bashlykov, Batra, Bhargava, Bhosale, Bikel, Blecher, Ferrer, Chen, Cucurull,
  Esiobu, Fernandes, Fu, Fu, Fuller, Gao, Goswami, Goyal, Hartshorn, Hosseini,
  Hou, Inan, Kardas, Kerkez, Khabsa, Kloumann, Korenev, Koura, Lachaux, Lavril,
  Lee, Liskovich, Lu, Mao, Martinet, Mihaylov, Mishra, Molybog, Nie, Poulton,
  Reizenstein, Rungta, Saladi, Schelten, Silva, Smith, Subramanian, Tan, Tang,
  Taylor, Williams, Kuan, Xu, Yan, Zarov, Zhang, Fan, Kambadur, Narang,
  Rodriguez, Stojnic, Edunov, and Scialom]{touvron2023llama2}
Touvron, H., Martin, L., Stone, K., Albert, P., Almahairi, A., Babaei, Y.,
  Bashlykov, N., Batra, S., Bhargava, P., Bhosale, S., Bikel, D., Blecher, L.,
  Ferrer, C.~C., Chen, M., Cucurull, G., Esiobu, D., Fernandes, J., Fu, J., Fu,
  W., Fuller, B., Gao, C., Goswami, V., Goyal, N., Hartshorn, A., Hosseini, S.,
  Hou, R., Inan, H., Kardas, M., Kerkez, V., Khabsa, M., Kloumann, I., Korenev,
  A., Koura, P.~S., Lachaux, M.-A., Lavril, T., Lee, J., Liskovich, D., Lu, Y.,
  Mao, Y., Martinet, X., Mihaylov, T., Mishra, P., Molybog, I., Nie, Y.,
  Poulton, A., Reizenstein, J., Rungta, R., Saladi, K., Schelten, A., Silva,
  R., Smith, E.~M., Subramanian, R., Tan, X.~E., Tang, B., Taylor, R.,
  Williams, A., Kuan, J.~X., Xu, P., Yan, Z., Zarov, I., Zhang, Y., Fan, A.,
  Kambadur, M., Narang, S., Rodriguez, A., Stojnic, R., Edunov, S., and
  Scialom, T.
\newblock Llama 2: Open foundation and fine-tuned chat models.
\newblock \emph{arXiv preprint arXiv:2307.09288}, 2023.

\bibitem[Trager et~al.(2023)Trager, Perera, Zancato, Achille, Bhatia, and
  Soatto]{trager2023linear}
Trager, M., Perera, P., Zancato, L., Achille, A., Bhatia, P., and Soatto, S.
\newblock Linear spaces of meanings: Compositional structures in
  vision-language models.
\newblock In \emph{Proceedings of the IEEE/CVF International Conference on
  Computer Vision}, pp.\  15395--15404, 2023.

\bibitem[Turner et~al.(2023)Turner, Thiergart, Udell, Leech, Mini, and
  MacDiarmid]{ActivationAddition:2023}
Turner, A.~M., Thiergart, L., Udell, D., Leech, G., Mini, U., and MacDiarmid,
  M.
\newblock Activation addition: Steering language models without optimization.
\newblock \emph{arXiv preprint arXiv:2308.10248}, art. arXiv:2308.10248, August
  2023.
\newblock \doi{10.48550/arXiv.2308.10248}.

\bibitem[Ushio et~al.(2021)Ushio, Anke, Schockaert, and
  Camacho-Collados]{ushio2021bert}
Ushio, A., Anke, L.~E., Schockaert, S., and Camacho-Collados, J.
\newblock {BERT is to NLP what AlexNet is to CV}: {Can} pre-trained language
  models identify analogies?
\newblock In \emph{Proceedings of the 59th Annual Meeting of the Association
  for Computational Linguistics and the 11th International Joint Conference on
  Natural Language Processing (Volume 1: Long Papers)}, pp.\  3609--3624, 2021.

\bibitem[Vaswani et~al.(2017)Vaswani, Shazeer, Parmar, Uszkoreit, Jones, Gomez,
  Kaiser, and Polosukhin]{vaswani2017attention}
Vaswani, A., Shazeer, N., Parmar, N., Uszkoreit, J., Jones, L., Gomez, A.~N.,
  Kaiser, {\L}., and Polosukhin, I.
\newblock Attention is all you need.
\newblock \emph{Advances in Neural Information Processing Systems}, 30, 2017.

\bibitem[Vylomova et~al.(2016)Vylomova, Rimell, Cohn, and
  Baldwin]{vylomova2016take}
Vylomova, E., Rimell, L., Cohn, T., and Baldwin, T.
\newblock Take and took, gaggle and goose, book and read: Evaluating the
  utility of vector differences for lexical relation learning.
\newblock In \emph{Proceedings of the 54th Annual Meeting of the Association
  for Computational Linguistics (Volume 1: Long Papers)}, pp.\  1671--1682,
  2016.

\bibitem[Wang et~al.(2023)Wang, Gui, Negrea, and Veitch]{wang2023concept}
Wang, Z., Gui, L., Negrea, J., and Veitch, V.
\newblock Concept algebra for score-based conditional models.
\newblock \emph{arXiv preprint arXiv:2302.03693}, 2023.

\bibitem[Zhu \& de~Melo(2020)Zhu and de~Melo]{zhu2020sentence}
Zhu, X. and de~Melo, G.
\newblock Sentence analogies: Linguistic regularities in sentence embeddings.
\newblock In \emph{Proceedings of the 28th International Conference on
  Computational Linguistics}, pp.\  3389--3400, 2020.

\bibitem[Zimmermann et~al.(2021)Zimmermann, Sharma, Schneider, Bethge, and
  Brendel]{zimmermann2021contrastive}
Zimmermann, R.~S., Sharma, Y., Schneider, S., Bethge, M., and Brendel, W.
\newblock Contrastive learning inverts the data generating process.
\newblock In \emph{International Conference on Machine Learning}, pp.\
  12979--12990. PMLR, 2021.

\bibitem[Zou et~al.(2023)Zou, Phan, Chen, Campbell, Guo, Ren, Pan, Yin,
  Mazeika, Dombrowski, Goel, Li, Byun, Wang, Mallen, Basart, Koyejo, Song,
  Fredrikson, Kolter, and Hendrycks]{zou2023representation}
Zou, A., Phan, L., Chen, S., Campbell, J., Guo, P., Ren, R., Pan, A., Yin, X.,
  Mazeika, M., Dombrowski, A.-K., Goel, S., Li, N., Byun, M.~J., Wang, Z.,
  Mallen, A., Basart, S., Koyejo, S., Song, D., Fredrikson, M., Kolter, Z., and
  Hendrycks, D.
\newblock Representation engineering: A top-down approach to {AI} transparency.
\newblock \emph{arXiv preprint arXiv:2310.01405}, 2023.

\end{thebibliography}
\bibliographystyle{icml2024}

\newpage
\appendix
\onecolumn

\section{Summary of Main Results}\label{sec:summary}
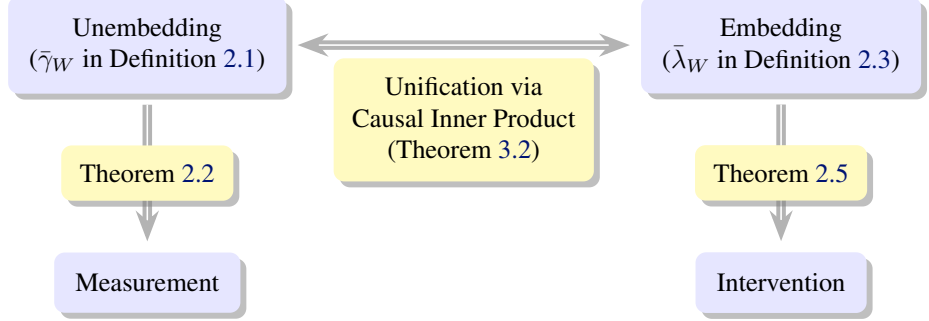
\begin{figure}[h]
    \centering
    \begin{tikzpicture}[
        mainnode/.style={rounded corners, fill=blue!10, drop shadow, align=center,  inner sep=8pt},
        arrownode/.style={rounded corners, drop shadow,  inner sep=7pt, align=center},
        arrow/.style={<->, double, >=Stealth, thick, draw=black!30, line width=1.5pt, bend angle=45, bend left, shorten >=4pt, shorten <=4pt}
    ]
    
        \node[mainnode] (emb) {Embedding \\ ($\bar\lambda_W$ in \cref{def:embedding_rep})};
        \node[mainnode] (unemb) [left=4.7cm of emb] {Unembedding \\ ($\bar\gamma_W$ in \cref{def:unembedding_rep})};
        \node[mainnode] (intervention) [below=2.1cm of emb] {Intervention};
        \node[mainnode] (measurement) [below=2.1cm of unemb] {Measurement};
        
        \draw[arrow, ->] (emb) -- node[arrownode,  fill=yellow!30] {\cref{thm:intervention}} (intervention);
        \draw[arrow, ->] (unemb) -- node[arrownode,  fill=yellow!30] {\cref{thm:measurement}} (measurement);
        \draw[arrow, <->] (emb) -- node[arrownode, below = 0.15cm, fill=yellow!30] {Unification via \\ Causal Inner Product \\ (\cref{thm:isomorphism})} (unemb);
    \end{tikzpicture}
    \caption{A high-level summary of our main results, illustrating the connections between the different notions of linear representations.}
    \label{fig:connection}
\end{figure}

In \Cref{fig:connection}, we give a high-level summary of our main results.
In \Cref{sec:linear_rep_hyp}, we have given the definitions of unembedding and embedding representations and how they also  yield measurement and intervention representations, respectively.
In \Cref{sec:inner_product}, we have defined the causal inner product and show how it unifies the unembedding and embedding representations via the induced Riesz isomorphism.

\section{Proofs}\label{sec:proofs}

\subsection{Proof of \cref{thm:measurement}}
\measRep*
\begin{proof}
    The proof involves writing out the softmax sampling distribution and invoking \cref{def:unembedding_rep}.
    \begin{align}
        &\logit \Pr(Y=Y(1) \given Y\in\{Y(0), Y(1)\}, \lambda) \\
        &= \log \frac{ \Pr(Y=Y(1) \mid Y \in \{Y(0), Y(1)\}, \lambda) }{ \Pr(Y=Y(0) \mid Y \in \{Y(0), Y(1)\}, \lambda) } \\
        &= \lambda^\top \left\{ \gamma(Y(1)) - \gamma(Y(0)) \right\} \label{eqn:linrep} \\
        &= \alpha \cdot \lambda^\top \bar\gamma_W. \label{eqn:affspan}
    \end{align}
    In~\eqref{eqn:linrep}, we simply write out the softmax distribution, allowing us to cancel out the normalizing constants for the two probabilities. 
    Equation~\eqref{eqn:affspan} follows directly from \cref{def:unembedding_rep}; note that the randomness of $\alpha$ comes from the randomness of $\{Y(0), Y(1)\}$.
\end{proof}

\subsection{Proof of \cref{lem:unembedding-embedding}}
\UnembeddingEmbedding*
\begin{proof}
    Let $\lambda_0,\lambda_1$ be a pair of embeddings such that
    \begin{equation}\label{eq:cond_embed_rep}
        \frac{\Pr(W=1 \given \lambda_1)}{\Pr(W=1\given \lambda_0)}>1 \quad\text{and}\quad \frac{\Pr(W,Z \given \lambda_1)}{\Pr(W, Z \given \lambda_0)} = \frac{\Pr(W \given \lambda_1)}{\Pr(W \given \lambda_0)},
    \end{equation}
    for any concept $Z$ that is causally separable with $W$. Then, by \cref{def:embedding_rep},
    \begin{equation}
        \lambda_1 - \lambda_0\in \Cone{\bar{\lambda}_W}.
    \end{equation}
    The condition \eqref{eq:cond_embed_rep} is equivalent to
    \begin{align}
        \frac{\Pr(W=1 \given \lambda_1)}{\Pr(W = 1 \given \lambda_0)} > 1 \quad \text{and}\quad \frac{\Pr(Z = 1 \given W,\lambda_1)}{\Pr(Z = 1 \given W, \lambda_0)} = 1. \label{relationship}
    \end{align}
    These two conditions are also equivalent to the following pair of conditions, respectively:
    \begin{equation}\label{eq:lemma_cond_pos}
        \frac{\Pr(Y=Y(1) \given Y\in\{Y(0), Y(1)\}, \lambda_1)}{\Pr(Y=Y(1) \given Y\in\{Y(0), Y(1)\},\lambda_0)} > 1
    \end{equation}
    and
    \begin{equation}\label{eq:lemma_cond_zero}
        \frac{\Pr(Y=Y(W, 1) \given Y\in\{Y(W,0), Y(W,1)\}, \lambda_1)}{\Pr(Y=Y(W, 1) \given Y\in\{Y(W,0), Y(W,1)\}, \lambda_0)} = 1
    \end{equation}
    The reason is that, conditional on $Y \in \{Y(0,0), Y(0, 1), Y(1, 0), Y(1,1)\}$, conditioning on $W$ is equivalent to conditioning on $Y \in \{Y(W,0), Y(W,1)\}$. And, the event $Z=1$ is equivalent to the event $Y=Y(W,1)$.
    (In words: if we know the output is one of ``king'', ``queen'', ``roi'', ``reine'' then conditioning on $W=1$ is equivalent to conditioning on the output being ``king'' or ``roi''. Then, predicting whether the word is in English is equivalent to predicting whether the word is ``king''.)
    
    By \cref{thm:measurement}, the two conditions \eqref{eq:lemma_cond_pos} and \eqref{eq:lemma_cond_zero} are respectively equivalent to
    \begin{equation}\label{eq:lemma_cond_alpha}
        \alpha(Y(0), Y(1)) (\lambda_1 - \lambda_0)^\top \bar\gamma_W > 0 \quad \text{and}\quad \alpha(Y(W,0), Y(W,1)) (\lambda_1 - \lambda_0)^\top \bar\gamma_Z = 0,
    \end{equation}
    where $\alpha$'s are positive a.s.
    These are in turn respectively equivalent to
    \begin{equation}\label{eq:cond_lemma}
        \bar\lambda_W^\top \bar\gamma_W >0 \quad\text{and}\quad \bar\lambda_W^\top \bar\gamma_Z = 0 .
    \end{equation}
    
    Conversely, if a representation $\bar\lambda_W$ satisfies \eqref{eq:cond_lemma} and there exist concepts $\{Z_i\}_{i=1}^{d-1}$ such that each concept is causally separable with $W$ and $\{\bar\gamma_W\}\cup\{\bar\gamma_{Z_i}\}_{i=1}^{d-1}$ is the basis of $\Reals^d$, then $\bar\lambda_W$ is unique up to positive scaling. If there exists $\lambda_0$ and $\lambda_1$ satisfying \eqref{eq:cond_embed_rep}, then the equivalence between \eqref{eq:cond_embed_rep} and \eqref{eq:lemma_cond_alpha} says that
    \begin{equation}
        (\lambda_1 - \lambda_0)^\top \bar\gamma_W > 0 \quad \text{and}\quad (\lambda_1 - \lambda_0)^\top \bar\gamma_Z = 0.
    \end{equation}
    In other words, $\lambda_1 - \lambda_0$ also satisfies \eqref{eq:cond_lemma}, implying that it must be the same as $\bar\lambda_W$ up to positive scaling.
    Therefore, for any $\lambda_0$ and $\lambda_1$ satisfying \eqref{eq:cond_embed_rep}, $\lambda_1 - \lambda_0 \in \Cone{\bar{\lambda}_W}$.
\end{proof}

\subsection{Proof of \cref{thm:intervention}}
\intervention*
\begin{proof}
    By \cref{thm:measurement},
    \begin{align}
        &\logit\Pr(Y= Y(W,1) \given Y\in\{Y(W,0), Y(W,1)\}, \lambda + c\bar\lambda_W)\\
        &= \alpha \cdot (\lambda+c\bar\lambda_W)^\top \bar\gamma_Z\\
        &=\alpha \cdot \lambda^\top \bar\gamma_Z + \alpha c \cdot \bar\lambda_W^\top \bar\gamma_Z
    \end{align}
    Therefore, the first probability is constant since $\bar\lambda_W^\top \bar\gamma_Z =0$ by \cref{lem:unembedding-embedding}.
    
    Also, by \cref{thm:measurement},
    \begin{align}
        &\logit\Pr(Y = Y(1,Z) \given Y\in\{Y(0,Z), Y(1,Z)\}, \lambda + c\bar\lambda_W)\\
        &= \alpha \cdot (\lambda+c\bar\lambda_W)^\top \bar\gamma_W\\
        &=\alpha \cdot \lambda^\top \bar\gamma_Z + \alpha c \cdot \bar\lambda_W^\top \bar\gamma_W
    \end{align}
    Therefore, the second probability is increasing since $\bar\lambda_W^\top \bar\gamma_W >0$ by \cref{lem:unembedding-embedding}.
\end{proof}

\subsection{Proof of \cref{thm:isomorphism}}
\RepresentationUnification*
\begin{proof}
    The causal inner product defines the Riesz isomorphism $\phi$ such that $\phi(\bar\gamma) = \ip{\bar\gamma}{\cdot}_{\mathrm{C}}$. Then, we have
    \begin{equation}
        \phi(\bar\gamma_W)(\bar\gamma_W) = \ip{\bar\gamma_W}{\bar\gamma_W}_{\mathrm{C}} >0\quad \text{and}\quad \phi(\bar\gamma_W)(\bar\gamma_Z) = \ip{\bar\gamma_W}{\bar\gamma_Z}_{\mathrm{C}} =0 ,
    \end{equation}
    where the second equality follows from \cref{def:causal_inner_product}.
    By \cref{lem:unembedding-embedding}, $\phi(\bar{\gamma}_W)$ expresses the unique unembedding representation $\bar\lambda_W$ (up to positive scaling); specifically, $\phi(\bar\gamma_W) = \bar\lambda_W^\top$ where $\bar\lambda_W^\top: \bar\gamma \mapsto \bar\lambda_W^\top\bar\gamma$.
\end{proof}

\subsection{Proof of \cref{thm:explicit_form}}
\ExplicitCIP*
\begin{proof}
    Since $\ip{\cdot}{\cdot}_{\mathrm{C}}$ is a causal inner product,
    \begin{equation}
        0 =  \bar\gamma_W^\top M \bar\gamma_Z\label{eq:first}
    \end{equation}
    for any causally separable concepts $W$ and $Z$. 
    By applying \eqref{eq:first} to the canonical representations $G = [\bar{\gamma}_{W_1}, \cdots, \bar{\gamma}_{W_d}]$, we obtain
    \begin{equation}
        I = G^\top M G.
    \end{equation}
    This shows that $M = G^{-\top} G^{-1}$, proving the first half of \eqref{eq:diagonalized}.
    
    Next, observe that $M \bar\gamma_{W_i}$ is an embedding representation for each concept $W_i$ for $i=1,\cdots, d$ by the proof of \cref{lem:unembedding-embedding} and \cref{thm:isomorphism}. Then, by \cref{ass:ortho_to_indep},
    \begin{align}
        0 &= \cov(\bar{\gamma}_{W_i}^\top M \gamma, \bar{\gamma}_{W_j}^\top M \gamma) \\
          &= \bar{\gamma}_{W_i}^\top M \cov(\gamma) M \bar{\gamma}_{W_j}.
    \end{align}
    for $i \neq j$. 
    Thus,
    \begin{equation}
        D^{-1} = G^\top M \cov (\gamma) M G,
    \end{equation}
    for some diagonal matrix $D$ with positive entries. Substituting in $M = G^{-\top} G^{-1}$, we get
    \begin{equation}
        \cov (\gamma) = G D^{-1}G^{\top},
    \end{equation}
    proving the second half of \eqref{eq:diagonalized}.
\end{proof}

\section{Experiment Details}\label{sec:exp_details}

\paragraph{The LLaMA-2 model}
We utilize the \texttt{llama-2-7b} variant of the LLaMA-2 model~\citep{touvron2023llama2}, which is accessible online (with permission) via the \texttt{huggingface} library.\footnote{\url{https://huggingface.co/meta-llama/Llama-2-7b-hf}}
Its seven billion parameters are pre-trained on two trillion \texttt{sentencepiece}~\citep{kudo2018sentencepiece} tokens, 90\% of which is in English. 
This model uses 32,000 tokens and 4,096 dimensions for its token embeddings.

\begin{table}[t]
\caption{Concept names, one example of the counterfactual pairs, and the number of the used pairs}
\label{tbl:concept_names}
\vskip 0.15in
\begin{center}
\begin{small}
\begin{tabular}{cccc}
\toprule
\# & Concept & Example & Count \\
\midrule
1 & verb $\Rightarrow$ 3pSg & (accept, accepts) & 32 \\
2 & verb $\Rightarrow$ Ving & (add, adding) & 31 \\
3 & verb $\Rightarrow$ Ved & (accept, accepted) & 47 \\
4 & Ving $\Rightarrow$ 3pSg & (adding, adds) & 27 \\
5 & Ving $\Rightarrow$ Ved & (adding, added) & 34 \\
6 & 3pSg $\Rightarrow$ Ved & (adds, added) & 29 \\
7 & verb $\Rightarrow$ V + able & (accept, acceptable) & 6 \\
8 & verb $\Rightarrow$ V + er & (begin, beginner) & 14 \\
9 & verb $\Rightarrow$ V + tion & (compile, compilation) & 8 \\
10 & verb $\Rightarrow$ V + ment & (agree, agreement) & 11 \\
11 & adj $\Rightarrow$ un + adj & (able, unable) & 5 \\
12 & adj $\Rightarrow$ adj + ly & (according, accordingly) & 18 \\
13 & small $\Rightarrow$ big & (brief, long) & 20 \\
14 & thing $\Rightarrow$ color & (ant, black) & 21 \\
15 & thing $\Rightarrow$ part & (bus, seats) & 13 \\
16 & country $\Rightarrow$ capital & (Austria, Vienna) & 15 \\
17 & pronoun $\Rightarrow$ possessive & (he, his) & 4 \\
18 & male $\Rightarrow$ female & (actor, actress) & 11 \\
19 & lower $\Rightarrow$ upper & (always, Always) & 34 \\
20 & noun $\Rightarrow$ plural & (album, albums) & 63 \\
21 & adj $\Rightarrow$ comparative & (bad, worse) & 19 \\
22 & adj $\Rightarrow$ superlative & (bad, worst) & 9 \\
23 & frequent $\Rightarrow$ infrequent & (bad, terrible) & 32 \\
24 & English $\Rightarrow$ French & (April, avril) & 46 \\
25 & French $\Rightarrow$ German & (ami, Freund) & 35 \\
26 & French $\Rightarrow$ Spanish & (année, año) & 35 \\
27 & German $\Rightarrow$ Spanish & (Arbeit, trabajo) & 22 \\
\bottomrule
\end{tabular}
\end{small}
\end{center}
\vskip -0.1in
\end{table}

\paragraph{Counterfactual pairs}
Tokenization poses a challenge in using certain words.
First, a word can be tokenized to more than one token. For example, a word ``princess'' is tokenized to ``prin'' + ``cess'', and $\gamma(\text{``princess''})$ does not exist. Thus, we cannot obtain the meaning of the exact word ``princess". Second, a word can be used as one of the tokens for another word. For example, the French words ``bas'' and ``est'' (``down'' and ``east'' in English) are in the tokens for the words ``basalt'', ``baseline'', ``basil'', ``basilica'', ``basin'', ``estuary'', ``estrange'', ``estoppel'', ``estival'', ``esthetics'', and ``estrogen''. Therefore, a word can have another meaning other than the meaning of the exact word.

When we collect the counterfactual pairs to identify $\bar\gamma_W$, the first issue in the pair can be handled by not using it. However, the second issue cannot be handled, and it gives a lot of noise to our results. \Cref{tbl:concept_names} presents the number of the counterfactual pairs for each concept and one example of the pairs. The pairs for 13, 17, 19, 23-27th concepts are generated by ChatGPT-4~\citep{openai2023gpt4}, and those for 16th concept are based on the csv file\footnote{\url{https://github.com/jmerullo/lm_vector_arithmetic/blob/main/world_capitals.csv}}). The other concepts are based on The Bigger Analogy Test Set (BATS)~\citep{gladkova2016analogy}, version 3.0\footnote{\url{https://vecto.space/projects/BATS/}}, which is used for evaluation of the word analogy task.

\paragraph{Context samples}
In \Cref{sec:exp_measurement}, for a concept $W$ (e.g., \ConceptDirName{English}{French}), we choose several counterfactual pairs $(Y(0), Y(1))$ (e.g., (house, maison)), then sample context $\{x_j^0\}$ and $\{x_j^1\}$ that the next token is $Y(0)$ and $Y(1)$, respectively, from Wikipedia. 
These next token pairs are collected from the \texttt{word2word} bilingual lexicon~\citep{choe2020word2word}, which is a publicly available word translation dictionary.
We take all word pairs between languages that are the top-1 correspondences to each other in the bilingual lexicon and filter out pairs that are single tokens in the LLaMA-2 model's vocabulary.

\Cref{tbl:measurement_pairs} presents the number of the contexts $\{x_j^0\}$ and $\{x_j^1\}$ for each concept and one example of the pairs $(Y(0), Y(1))$.
\begin{table}[t]
\caption{Concepts used to investigate measurement notion}
\label{tbl:measurement_pairs}
\vskip 0.15in
\begin{center}
\begin{small}
\begin{tabular}{ccc}
\toprule
Concept & Example & Count \\
\midrule
English $\Rightarrow$ French & (house, maison) & (209, 231) \\
French $\Rightarrow$ German & (déjà, bereits) & (278, 205) \\
French $\Rightarrow$ Spanish & (musique, música) & (218, 214) \\
German $\Rightarrow$ Spanish & (Krieg, guerra) & (214, 213) \\
\bottomrule
\end{tabular}
\end{small}
\end{center}
\vskip -0.1in
\end{table}

In the experiment for intervention notion, for a concept $W, Z$, we sample texts which $Y(0,0)$ (e.g., ``king'') should follow, via ChatGPT-4. We discard the contexts such that $Y(0,0)$ is not the top 1 next word. \Cref{tbl:intervention_texts} present the contexts we use.
\begin{table}[t]
\caption{Contexts used to investigate intervention notion}
\label{tbl:intervention_texts}
\vskip 0.15in
\begin{center}
\begin{small}
\begin{tabular}{cc}
        \toprule
        $j$ & $x_j$\\
        \midrule
        1 & Long live the\\
        2 & The lion is the\\
        3 & In the hierarchy of medieval society, the highest rank was the\\
        4 & Arthur was a legendary\\
        5 & He was known as the warrior\\
        6 & In a monarchy, the ruler is usually a\\
        7 & He sat on the throne, the\\
        8 & A sovereign ruler in a monarchy is often a\\
        9 & His domain was vast, for he was a\\
        10 & The lion, in many cultures, is considered the\\
        11 & He wore a crown, signifying he was the\\
        12 & A male sovereign who reigns over a kingdom is a\\
        13 & Every kingdom has its ruler, typically a\\
        14 & The prince matured and eventually became the\\
        15 & In the deck of cards, alongside the queen is the\\
        \bottomrule
\end{tabular}
\end{small}
\end{center}
\vskip -0.1in
\end{table}

\section{Additional Results}\label{sec:additional_results}
\subsection{Histograms of random and counterfactual pairs for all concepts}\label{sec:additional_histograms}

\begin{figure}[t]
    \centering
    \includegraphics[width= 1.0\linewidth]{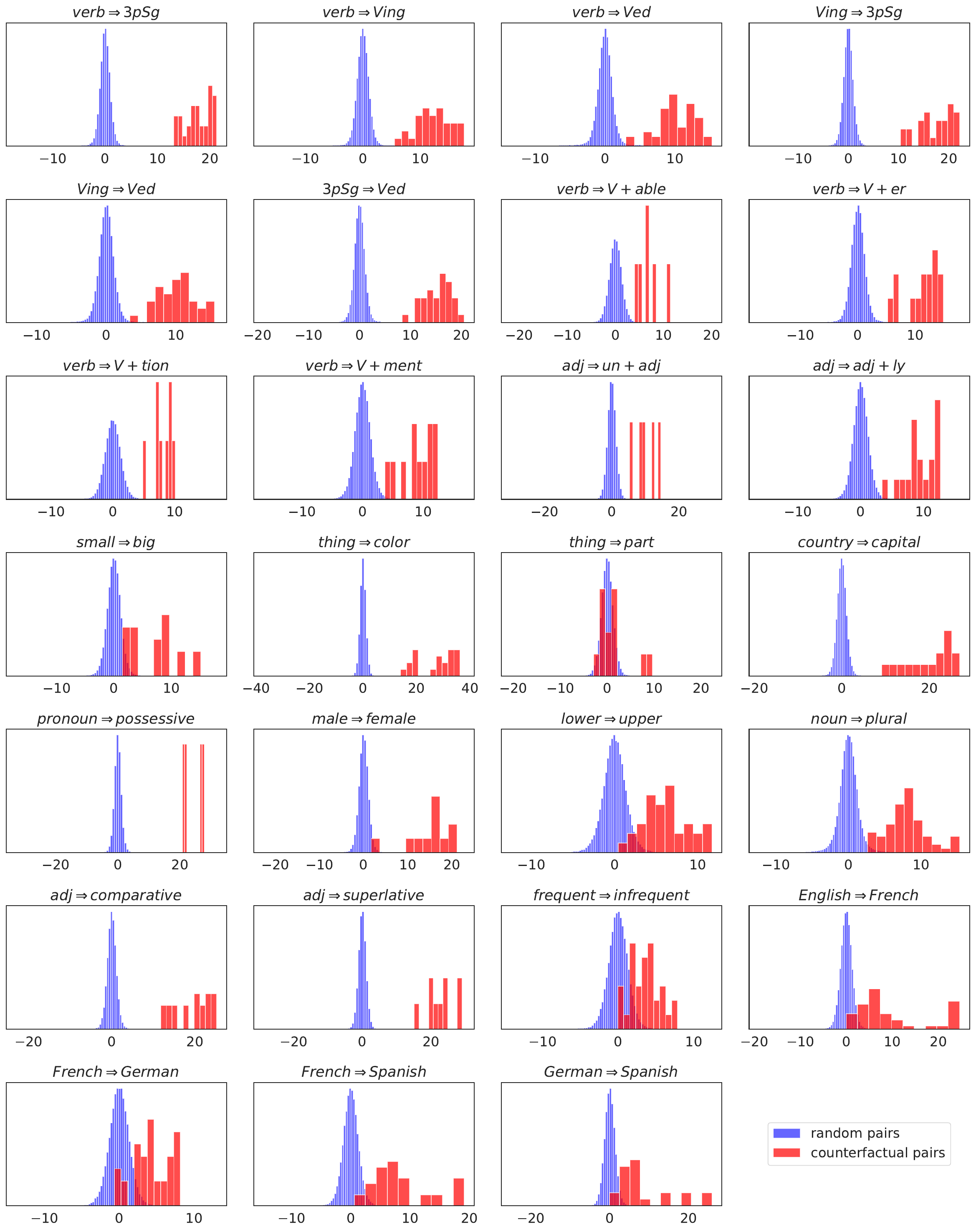}
    \caption{Histograms of the projections of the counterfactual pairs $\ip{\bar{\gamma}_{W, (-i)}}{\gamma(y_i(1)) - \gamma(y_i(0))}_{\mathrm{C}}$ (red), and the projections of the differences between 100K randomly sampled word pairs onto the estimated concept direction (blue).
    See \Cref{tbl:concept_names} for details about each concept $W$ (the title of each plot).}
    \label{fig:appendix_right-skewed}
\end{figure}

In \Cref{fig:appendix_right-skewed}, we include an analog of \Cref{fig:right-skewed} where we check the causal inner product of the differences between the counterfactual pairs and an LOO estimated unembedding representation for each of the 27 concepts. While the most of the concepts are encoded in the unembedding representation, some concepts, such as \ConceptDirName{thing}{part}, are not encoded in the unembedding space $\Gamma$.

\subsection{Comparison with the Euclidean inner products}\label{appendix:other_inner_product}
\begin{figure}[t]
    \centering
    \includegraphics[width=1.0\linewidth]{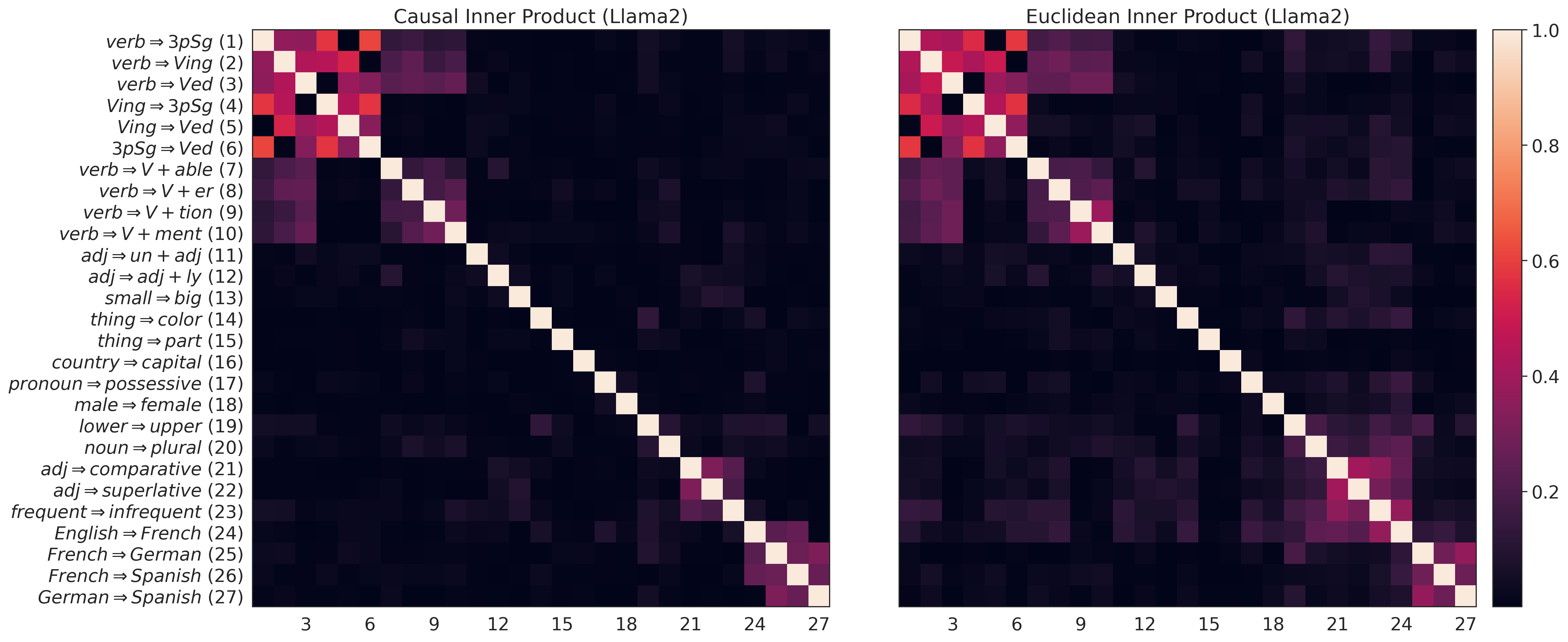}
    \caption{For the LLaMA-2-7B model, causally separable concepts are approximately orthogonal under the estimated causal inner product and, surprisingly, under the Euclidean inner product as well.
    The heatmaps show $ |\ip{\bar\gamma_W}{\bar\gamma_Z}|$ for the estimated unembedding representations of each concept pair $(W, Z)$.
    The plot on the left shows the estimated inner product based on~\eqref{eq:our_CIP}, and the right plot represents the Euclidean inner product.
    The detail for the concepts is given in \Cref{tbl:concept_names}.} 
    \label{fig:two_heatmaps_Llama2}
\end{figure}

\begin{figure}[t]
    \centering
    \includegraphics[width=1.0\linewidth]{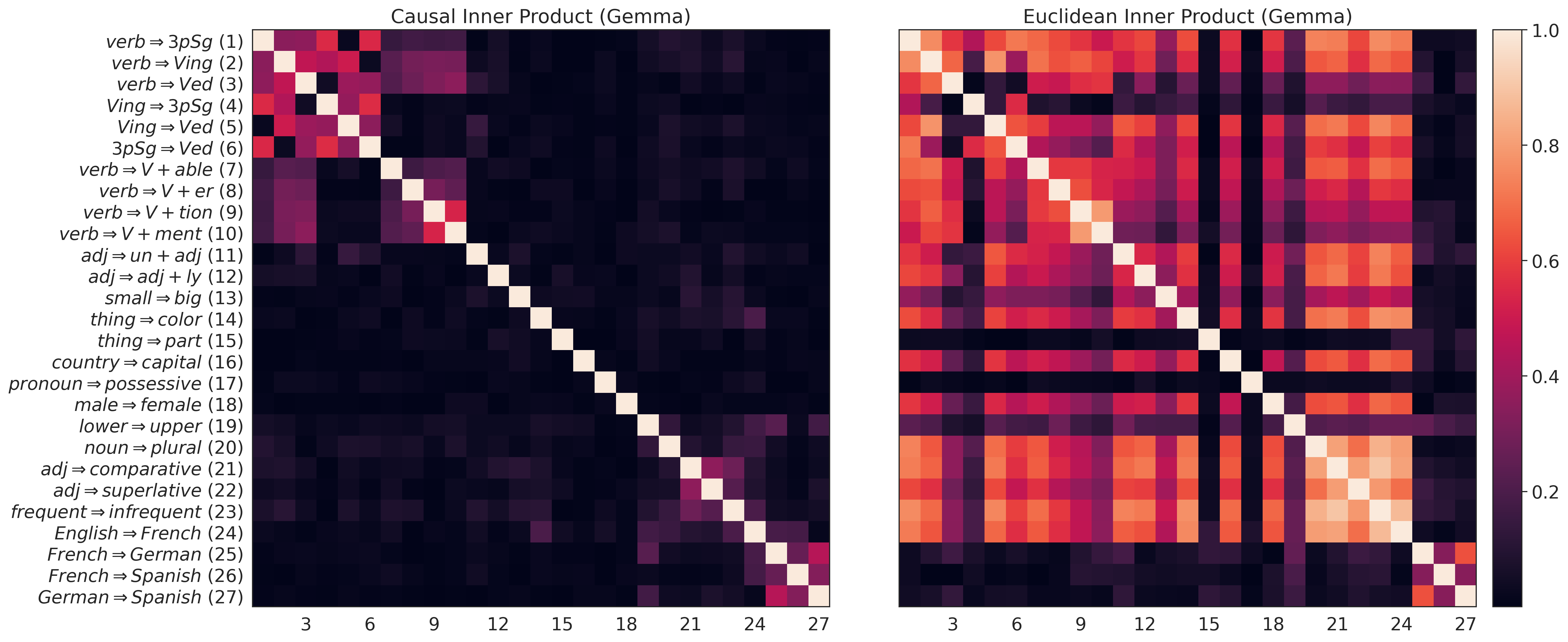}
    \caption{For the Gemma-2B model, causally separable concepts are approximately orthogonal under the estimated causal inner product; however, the Euclidean inner product does not capture semantics.
    The heatmaps show $ |\ip{\bar\gamma_W}{\bar\gamma_Z}|$ for the estimated unembedding representations of each concept pair $(W, Z)$.
    The plot on the left shows the estimated inner product based on~\eqref{eq:our_CIP}, and the right plot represents the Euclidean inner product.
    The detail for the concepts is given in \Cref{tbl:concept_names}.} 
    \label{fig:two_heatmaps_Gemma}
\end{figure}

In \cref{fig:two_heatmaps_Llama2}, we also plot the cosine similarities induced by the Euclidean inner product between the unembedding representations.
Surprisingly, the Euclidean inner product somewhat works in the LLaMA-2 model as most of the causally separable concepts are orthogonal!
This may due to some initialization or implicit regularizing effect that favors learning unembeddings with approximately isotropic covariance. 
Nevertheless, the estimated causal inner product clearly improves on the Euclidean inner product. For example, \ConceptDirName{frequent}{infrequent} (concept 23) has high Euclidean inner product with many separable concepts, and these are much smaller for the causal inner product. Conversely, \ConceptDirName{English}{French} (24) has low Euclidean inner product with the other language concepts (25-27), but high causal inner product with \ConceptDirName{French}{German} and \ConceptDirName{French}{Spanish} (while being nearly orthogonal to \ConceptDirName{German}{Spanish}, which does not share French).

Interestingly, the same heatmaps for a more recent Gemma-2B model \citep{team2024gemma} in \cref{fig:two_heatmaps_Gemma} illustrate that the Euclidean inner product doesn't capture semantics, while the causal inner product still works.
One possible reason is that the origin of the unembeddings is meaningful as the Gemma model ties the unembeddings to the token embeddings used before the transformer layers.

\subsection{Additional results from the measurement experiment}\label{sec:additional_measurements}

\begin{figure}[t]
    \centering
    \includegraphics[width= 1.0\linewidth]{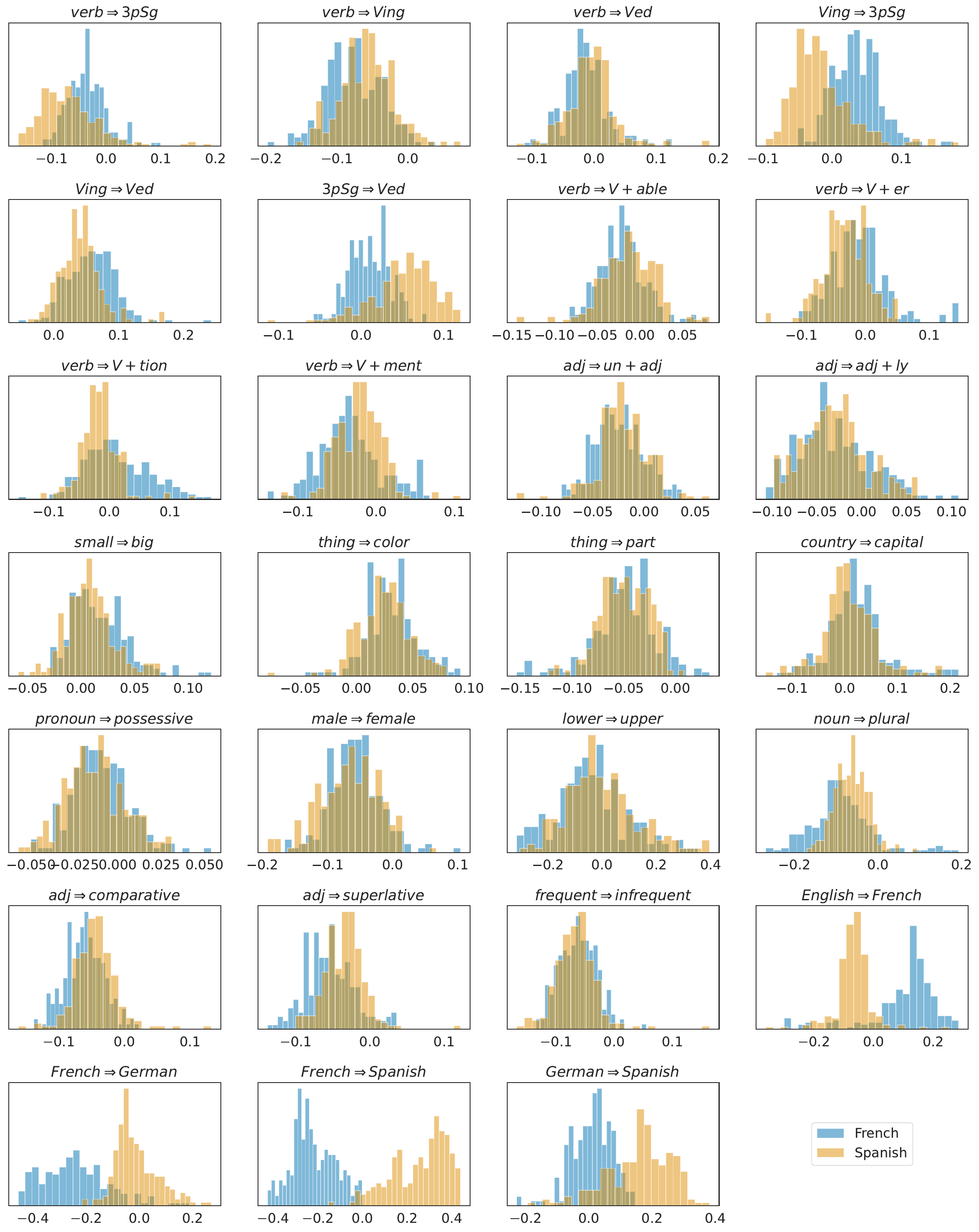}
    \caption{Histogram of $\bar\gamma_{C}^\top \lambda(x_j^\texttt{fr})$ vs  $\bar\gamma_{C}^\top \lambda(x_j^\texttt{es})$ for all concepts $C$, where $\{x_j^\texttt{fr}\}$ are random contexts from French Wikipedia, and $\{x_j^\texttt{es}\}$ are random contexts from Spanish Wikipedia.}
    \label{fig:appendix_measurement_French-Spanish}
\end{figure}

\begin{figure}[t]
    \centering
    \includegraphics[width= 1.0\linewidth]{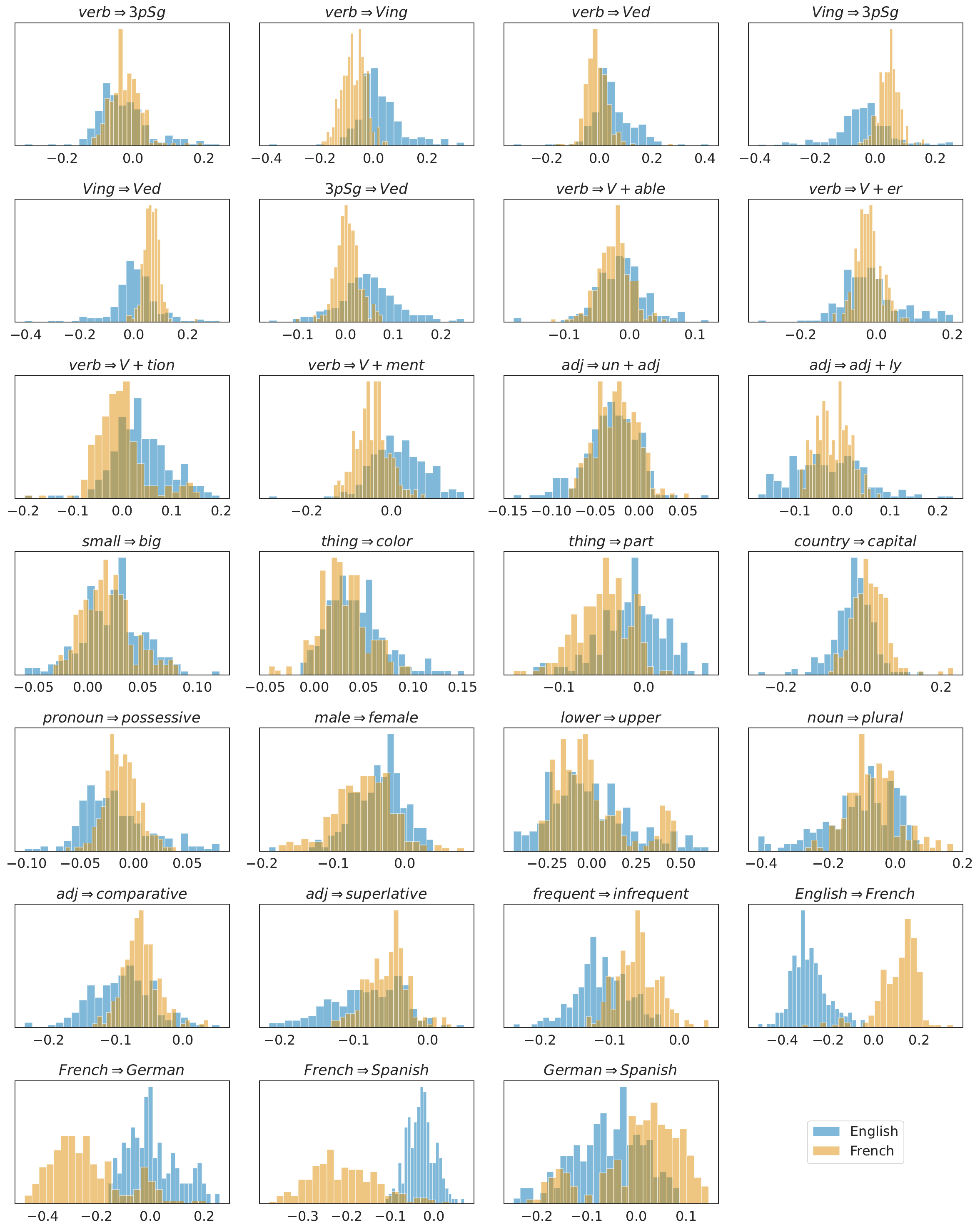}
    \caption{Histogram of $\bar\gamma_{C}^\top \lambda(x_j^\texttt{en})$ vs  $\bar\gamma_{C}^\top \lambda(x_j^\texttt{fr})$ for all concepts $C$, where $\{x_j^\texttt{en}\}$ are random contexts from English Wikipedia, and $\{x_j^\texttt{fr}\}$ are random contexts from French Wikipedia.}
    \label{fig:appendix_measurement_English-French}
\end{figure}

We include analogs of \Cref{fig:measurement}, specifically where we use each of the 27 concepts as a linear probe on either \ConceptDirName{French}{Spanish} (\Cref{fig:appendix_measurement_French-Spanish}) or \ConceptDirName{English}{French} (\Cref{fig:appendix_measurement_English-French}) contexts.

\subsection{Additional results from the intervention experiment}\label{sec:additional_intervention}

\begin{figure}[t]
    \centering
    \includegraphics[width= 1.0\linewidth]{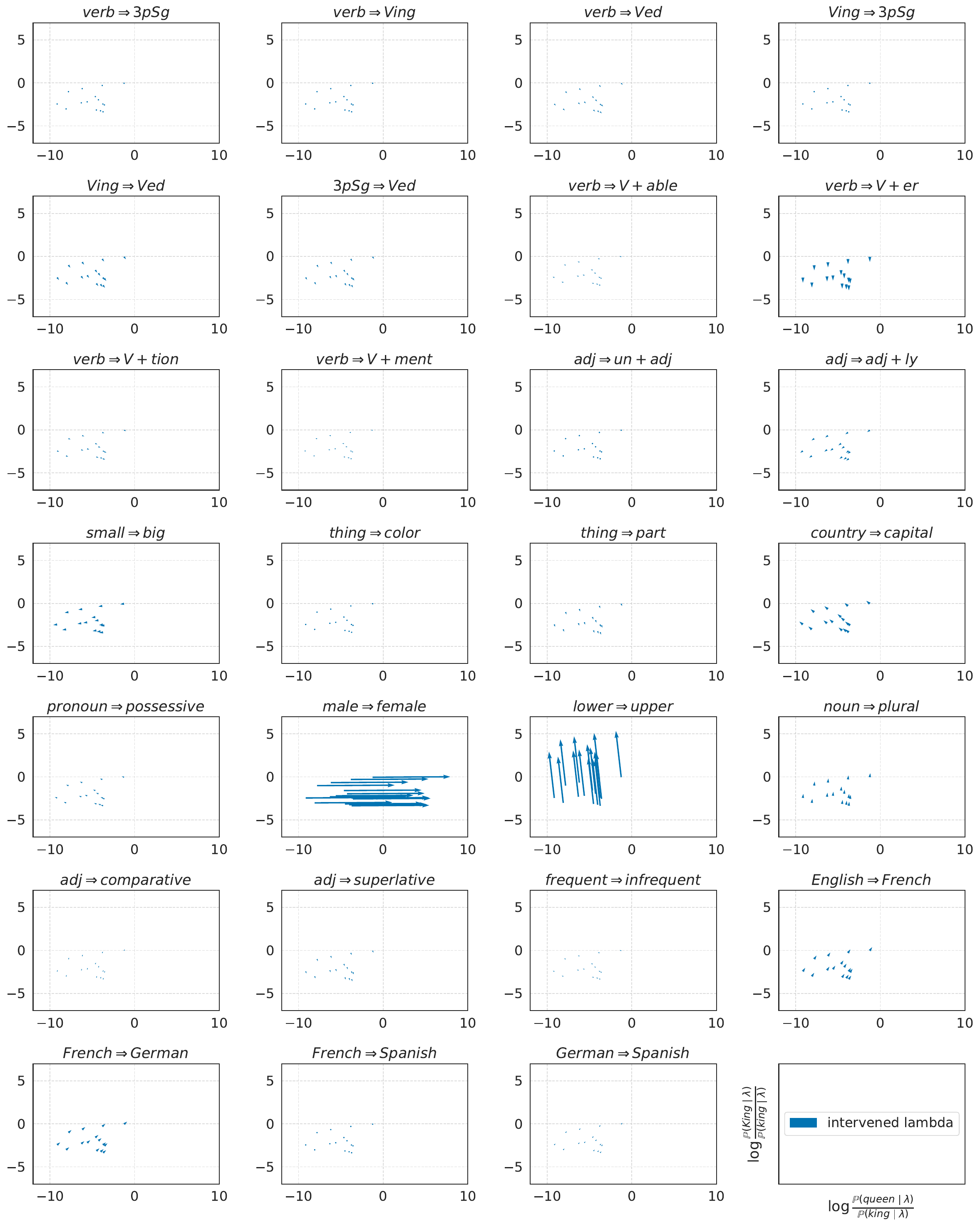}
    \caption{Change in $\log(\Pr(\text{``queen''}  \mid x ) / \Pr(\text{``king''}  \mid x ))$ and $\log(\Pr(\text{``King''}  \mid x ) / \Pr(\text{``king''}  \mid x ))$, after changing $\lambda(x_j)$ to $\lambda_{C,\alpha}(x_j)$ for $\alpha \in [0, 0.4]$ and any concept $C$. The starting point and ending point of each arrow correspond to the $\lambda(x_j)$ and $\lambda_{C,0.4}(x_j)$, respectively.}
    \label{fig:appendix_intervention_flip}
\end{figure}

In \Cref{fig:appendix_intervention_flip}, we include an analog of \Cref{fig:intervention_flip} where we add the embedding representation $\alpha \bar\lambda_C$ \eqref{eq:new_intervention_rep} for each of the 27 concepts to $\lambda(x_j)$ and see the change in logits.

\subsection{Additional tables of top-5 words after intervention}\label{sec:additional_tables}
\begin{table}[t]
\caption{Context: ``The prince matured and eventually became the ''}
\label{tbl:appendix_intervention_topk1}
\vskip 0.15in
\begin{center}
\begin{small}
\begin{tabular}{cccccc}
\toprule
Rank  & $\alpha = $ 0 & 0.1 & 0.2 & 0.3 & 0.4 \\
\midrule
1 & \it king & \it king & em & \bf queen & \bf queen \\
2 & em & em & r & em & woman \\
3 & leader & r & leader & r & lady \\
4 & r & leader & \it king & leader & wife \\
5 & King & head & \bf queen & woman & em \\
\bottomrule
\end{tabular}
\end{small}
\end{center}
\vskip -0.1in
\end{table}

\begin{table}[t]
\caption{Context: ``In a monarchy, the ruler is usually a ''}
\label{tbl:appendix_intervention_topk2}
\vskip 0.15in
\begin{center}
\begin{small}
\begin{tabular}{cccccc}
\toprule
Rank  & $\alpha = $ 0 & 0.1 & 0.2 & 0.3 & 0.4 \\
\midrule
1 & \it king & \it king & her & woman & woman \\
2 & monarch & monarch & monarch & \bf queen & \bf queen \\
3 & member & her & member & her & female \\
4 & her & member & woman & monarch & her \\
5 & person & person & \bf queen & member & member \\
\bottomrule
\end{tabular}
\end{small}
\end{center}
\vskip -0.1in
\end{table}
\Cref{tbl:appendix_intervention_topk1} and \Cref{tbl:appendix_intervention_topk2} are analogs of \Cref{tbl:intervention_topk} where we use different contexts $x=$ ``In a monarchy, the ruler usually is a '' and $x=$ ``The prince matured and eventually became the ''.
For the first example, note that ``r'' and ``em'' are the prefix tokens for words related to royalty, such as ``ruler'', ``royal'', and ``emperor''. For the second example, even when the target word ``\textbf{queen}'' does not become the most likely one, the most likely words still reflect the concept direction (``woman'', ``\textbf{queen}'', ``her'', ``female'').

\subsection{A sanity check for the estimated causal inner product}\label{appendix:check_assumption}
\begin{figure}[t]
    \centering
    \includegraphics[width= 1.0\linewidth]{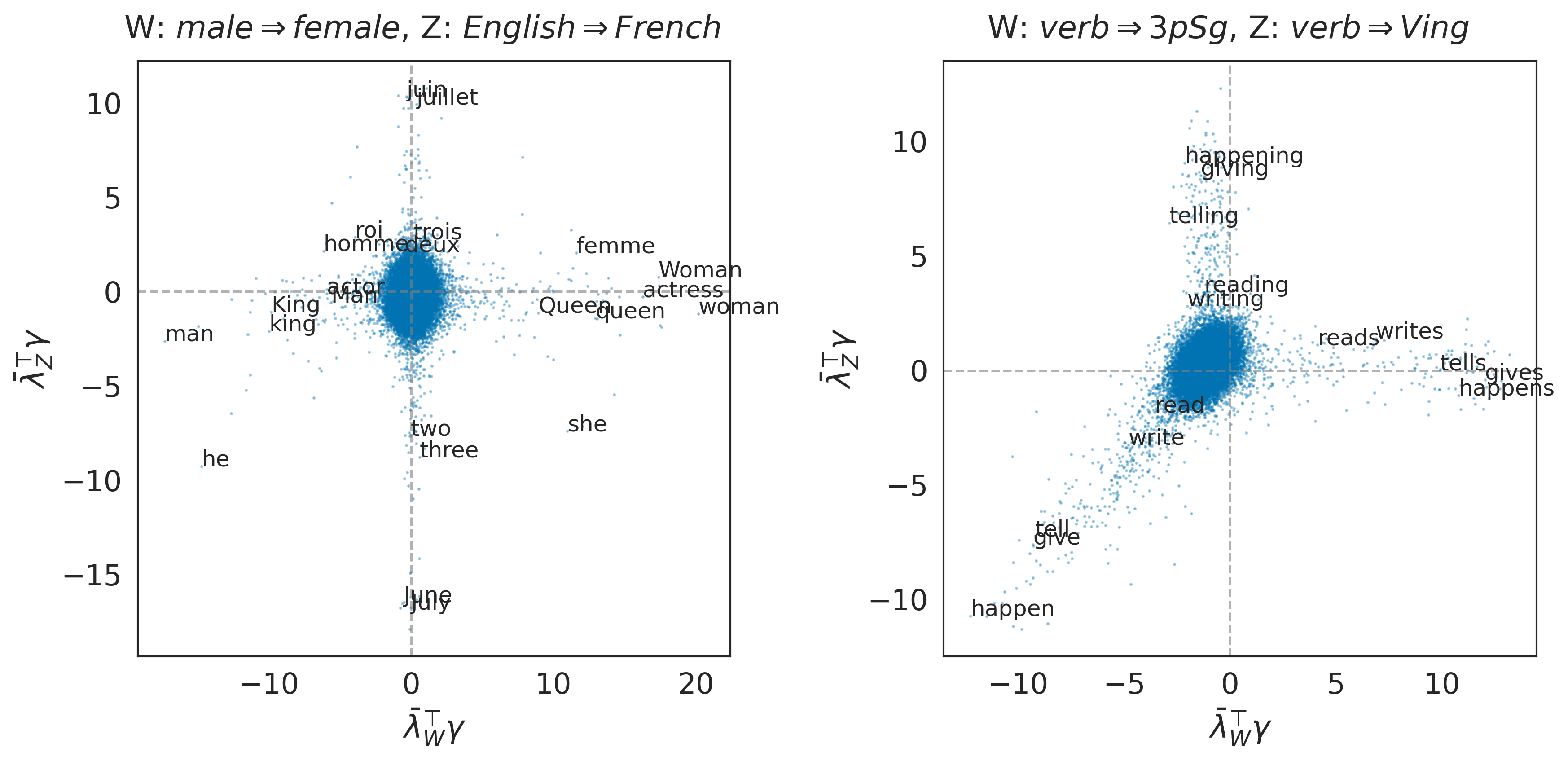}
    \caption{The left plot shows that $\bar\lambda_W^\top \gamma$ and $\bar\lambda_Z^\top \gamma$ are uncorrelated for the causally separable concepts $W=$ \ConceptDirName{male}{female} and $Z=$ \ConceptDirName{English}{French}.
    On the other hand, the right plot shows that $\bar\lambda_W^\top \gamma$ and $\bar\lambda_Z^\top \gamma$ are correlated for the non-causally separable concepts $W=$ \ConceptDirName{verb}{3pSg} and $Z=$ \ConceptDirName{verb}{Ving}.
    Each dot corresponds to the unembedding vector $\gamma$ for each token in the vocabulary.}
    \label{fig:appendix_independent}
\end{figure}

In earlier experiments, we found that the choice $M = \cov(\gamma)^{-1}$ from~\eqref{eq:our_CIP} yields a causal inner product and induces an embedding representation $\bar\lambda_W$ in the form of~\eqref{eq:new_intervention_rep}.
Here, we run a sanity check experiment where we verify that the induced embedding representation satisfies the uncorrelatedness condition in Assumption~\ref{appendix:check_assumption}.
In \Cref{fig:appendix_independent}, we empirically show that $\bar\lambda_W^\top \gamma$ and $\bar\lambda_Z^\top \gamma$ are uncorrelated for the causally separable concepts (left plot), while they are correlated for the non-causally separable concepts (right plot).
In these plots, each dot corresponds to the point $(\bar\lambda_W^\top \gamma, \bar\lambda_Z^\top \gamma)$, where $\gamma$ is an unembedding vector $\gamma$ corresponding to each token in the LLaMA-2 vocabulary (32K total).

\end{document}